\documentclass{article}

\usepackage[final]{corl_2026} 

\usepackage{amsmath}
\usepackage{amssymb}
\usepackage{bm}
\usepackage[ruled,vlined]{algorithm2e}
\usepackage{array}[=v2.4]
\usepackage{booktabs}
\usepackage{graphicx}
\usepackage{xcolor}
\newcommand{\hbc}[1]{#1}
\newenvironment{hbcrevision}{\par\begingroup\noindent}{\par\endgroup}
\usepackage{multirow}
\usepackage{enumitem}
\usepackage{makecell}
\usepackage{float}

\newcommand{\ours}{\textsc{wam-ttt}\xspace}
\usepackage{xspace}

\title{WAM-TTT:
Steering World-Action Models by Watching Human Play at Test Time}

\author{
  \normalfont
  \makebox[\textwidth][c]{%
    \normalfont\textbf{Yusen Feng}$^{1,2,*}$ \quad
    \textbf{Bingchen Han}$^{1,2,*}$ \quad
    \textbf{Jiangran Lyu}$^{1,2,*}$}\\
  \makebox[\textwidth][c]{%
    \normalfont Kai Liu$^{2,3}$ \,
    Yixin Zheng$^{2,3}$ \,
    Yuxuan Wan$^{1,2}$ \,
    Weiheng Liu$^{2,3}$ \,
    Sun Han$^{1,2}$ \,
    Ruiqin Li$^{1,2}$}\\
  \makebox[\textwidth][c]{%
    \normalfont Yulong Zhang$^{1}$ \,
    Fangfu Liu$^{4}$ \,
    Xuesong Shi$^{2}$}\\
  \makebox[\textwidth][c]{%
    \normalfont Yizhou Wang$^{1,\dagger}$ \,
    Libin Liu$^{1,\dagger}$ \,
    Zhizheng Zhang$^{2,\dagger}$ \,
    He Wang$^{1,2,\dagger}$}\\[0.5em]
  \normalfont $^{1}$Peking University \quad
  $^{2}$Galbot \quad
  $^{3}$CASIA \quad
  $^{4}$Tsinghua University\\
  \normalfont $^{*}$Equal contribution \quad
  $^{\dagger}$Corresponding authors
}

\begin{document}
\maketitle

\begin{figure}[h]
  \centering
  \includegraphics[width=0.80\linewidth]{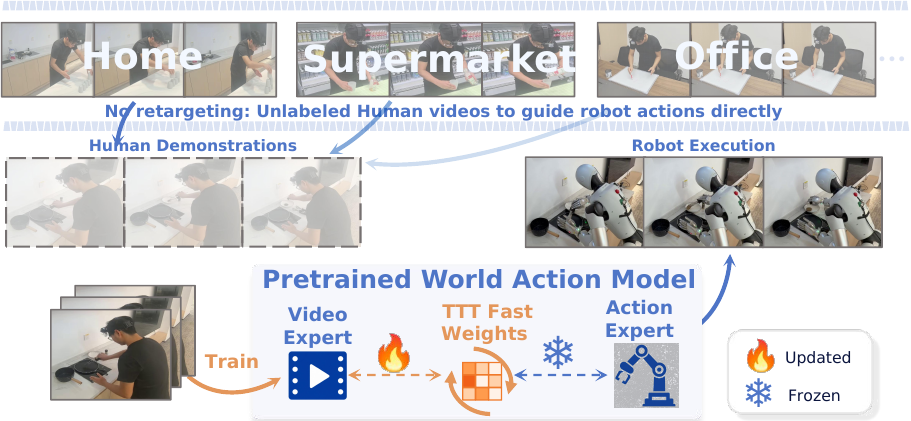}
  \caption{Overview of \ours. Given unlabeled human demonstrations from diverse environments, \ours steers a pretrained World Action Model (WAM) without retargeting, robot actions, or human-side annotations. During deployment, human videos are absorbed into lightweight TTT fast weights through self-supervised video prediction, while the pretrained action model remains frozen. The adapted memory then guides robot execution through the WAM's shared visual-action dynamics, enabling efficient and reusable steering from human demonstrations.}
  \label{fig:teaser}
\end{figure}


\begin{abstract}
Steering robot foundation models (RFMs) toward new task variants or user-preferred behaviors remains challenging, often requiring additional robot demonstrations, task-specific fine-tuning, or long-context conditioning. We present \ours, a test-time training framework for steering world action models  from raw human videos. Rather than treating human videos as trajectories to imitate, \ours absorbs them into a lightweight adaptive memory inside a frozen WAM through self-supervised video prediction. To make this memory useful for control, we introduce a meta-training stage that aligns human demonstrations with robot behaviors using paired human-robot data and a key--value memory reconstruction objective. At test time, only unlabeled human videos are required to adapt the memory, while the pretrained WAM remains frozen. This enables efficient and reusable steering without robot actions, human-side annotations, or task-specific fine-tuning, while preserving the generalization ability of the foundation model. Extensive experiments show that \ours consistently outperforms in-context human-video conditioning baselines across diverse manipulation tasks and generalization settings.
Our project website is available at \url{https://wam-ttt.github.io}.

\end{abstract}



\section{Introduction}
\label{sec:intro}

Recently, the robotics community has increasingly pursued general-purpose robot foundation models through large-scale pretraining.
However, most existing RFMs primarily absorb knowledge into fixed model parameters.
Once deployed, their behavior is largely determined by the pretrained weights and a limited conditioning interface, such as language instructions, goal images, or short observation histories\citep{yu2018daml, bahl2022whirl, xu2023xskill, bharadhwaj2024hopm, hansen2021pad, xu2024flowinterface}.
As a result, steering RFMs toward new task variants, object interactions, or user-preferred strategies typically requires collecting additional robot demonstrations or fine-tuning the full model.
This limits the flexibility and reusability of RFMs in open-ended deployment settings, where users may wish to quickly specify new behaviors without retraining a robot policy.

Human demonstrations offer a natural and scalable interface for steering RFMs\citep{kareer2024egomimic, hoque2025egodex, grauman2024egoexo4d, zheng2026egoscale, chen2025vidbot, kim2025uniskill}: users can simply show how objects should be handled, without specifying robot actions.
Existing methods typically leverage human videos through co-training or fine-tuning with robot data~\citep{kareer2024egomimic,hoque2025egodex,grauman2024egoexo4d,zheng2026egoscale,chen2025vividex,chen2025vidbot,kim2025uniskill}, often relying on additional supervision such as hand poses, 3D motion, or retargeted trajectories~\citep{bahl2022whirl,wang2023mimicplay,xu2023xskill,bharadhwaj2023h2r,bharadhwaj2024hopm,xu2024flowinterface,jain2024vid2robot,bharadhwaj2024gen2act}.
Such supervision can be noisy and costly to obtain, while task-specific fine-tuning may cause catastrophic forgetting and reduce the reusability of the pretrained model.
A more direct alternative is to condition robot policies on raw human videos~\citep{shah2025mimicdroid}, but this requires learning such capabilities during large-scale pretraining and incurs rapidly growing context lengths as demonstrations accumulate.

To address these challenges, we propose \ours, a test-time training framework for steering world action models (WAMs) with human demonstrations.
Rather than treating human videos as trajectories to imitate, \ours uses them as deployment-time memory learned through video prediction.
Since WAMs jointly model visual dynamics and actions, the adapted memory can steer action generation through the model's shared video-action representation.
To make this adaptation useful for control, we introduce a meta-training stage that aligns human demonstrations with robot behaviors using paired human-robot data and a key--value memory reconstruction loss.
At deployment, only human videos are required: the memory is updated through video prediction, while the pretrained WAM remains frozen.
As a result, \ours enables efficient and reusable steering of WAMs toward new task variants while preserving the generalization ability of the foundation model.

Extensive experiments show that \ours consistently outperforms in-context-learning-based human-video conditioning baselines.
Our ablations further demonstrate the importance of test-time memory adaptation, the video prediction objective, and the key--value memory reconstruction loss, highlighting the effectiveness of our design for steering pretrained WAMs.
Our contributions are threefold.
\begin{enumerate}
\item We formulate human-video-based steering of world action models as a test-time training problem, enabling deployment-time adaptation from raw human demonstrations without robot actions.

\item We propose \ours, a plug-and-play TTT memory that absorbs human videos into a frozen WAM through self-supervised video prediction, together with a human-robot alignment objective that makes the learned memory useful for robot control.

\item We demonstrate that \ours enables efficient and reusable steering from human demonstrations, outperforming in-context video conditioning while avoiding additional human-side annotations and full-model fine-tuning.

\end{enumerate}


\section{Related Work}
\label{sec:related}

\paragraph{World Action Models.}
World models have become an increasingly important interface for robot learning, as they provide a predictive substrate for reasoning about how actions change future observations\citep{chi2024eva,chi2025wow, Zhang2026QwenRobotWorldTR}. Recent world-action models (WAMs) go one step further by coupling future visual prediction with action generation, enabling policies to be grounded in imagined state transitions rather than in purely reactive action prediction~\citep{wang2023mimicplay,bharadhwaj2024gen2act,jain2024vid2robot,zhu2025unified,lda1b, bi2026motus, team2026motubrain, zhang2026imagewam, yu2026maskwam, bi2026motus, team2026motubrain, peng2026reworld, ye2026worldactionmodelszeroshot, ye2026gigaworld, lingbot-va2026, yuan2026fastwam, cai2026ahawam, harmowam2026, beingbeyond2026beingh07, guo2026xwam, lyu2025dywa, agarwal2026cosmos, ma2026dit4dit, physicalintelligence2026pi07, liu2026oawam, yang2026abot, chen2026abotm05}. This coupling also makes video a natural supervision signal: action-free videos can improve visual-dynamics representations, while robot trajectories can anchor those representations to executable actions~\citep{wang2023mimicplay,bharadhwaj2024gen2act,jain2024vid2robot,zhu2025unified, kim2025cosmospolicy, hu2024video}. However, existing world action models primarily focus on pretraining but ignoring the steeribility during deployment. 

\paragraph{Test-time training and adaptive memory.}
Test-time training adapts models using signals derived from test inputs, typically through self-supervised or entropy-based objectives under distribution shift~\citep{sun2020ttt,wang2021tent,liu2021tttpp,gandelsman2022tttmae, sun2025learnattesttime}. Recent work reframes this idea as memory: TTT layers and related fast-weight mechanisms store information from the current sequence in adaptive parameters rather than relying only on fixed activations or explicit KV caches~\citep{sun2025learnattesttime,behrouz2025titans,zhang2026spatialttt}. Robotics has also explored test-time adaptation through auxiliary losses, visual model-based objectives, or online environment feedback~\citep{hansen2021pad,yang2023movie,bai2025evolvevla,liu2026ttvla}. These methods typically adapt from robot observations, rewards, or interaction rollouts. \ours instead uses human demonstrations as the test-time information source. Rather than updating the full policy online, we calibrate a TTT branch during pre-training so that, at inference, human-video Key/Value features can act as a residual skill memory inside the WAM.

\paragraph{Learning from human video.}
Human videos provide scalable evidence about objects, contacts, and task progress, but transferring them to robots is difficult because human motion is not directly executable by a robot. Prior work addresses this gap by using human data for training-time representation learning, cross-domain alignment, or policy supervision~\citep{kareer2024egomimic,hoque2025egodex,grauman2024egoexo4d,zheng2026egoscale,chen2025vividex,chen2025vidbot,kim2025uniskill}. Other methods use human videos more directly, but often require hand pose, RGB-D motion, retargeted trajectories, object flow, latent plan extraction, generated videos, robot demonstrations, or online interaction~\citep{bahl2022whirl,wang2023mimicplay,xu2023xskill,bharadhwaj2023h2r,bharadhwaj2024hopm,xu2024flowinterface,jain2024vid2robot,bharadhwaj2024gen2act,shah2025mimicdroid}. In contrast, \ours treats a small set of unseen and unlabelled human play videos as deployment-time skill memory. The videos are not converted into robot actions or explicit human poses; instead, they provide Key/Value context to a calibrated TTT cross-attention branch, enabling skill transfer without retargeting, generated demonstrations, robot-context examples, interaction rollouts, or full deployment-time fine-tuning.

\section{Method}
\label{sec:method}

\subsection{Architecture}
\label{sec:method:architecture}

\begin{figure}[t]
    \centering
    \includegraphics[width=\linewidth]{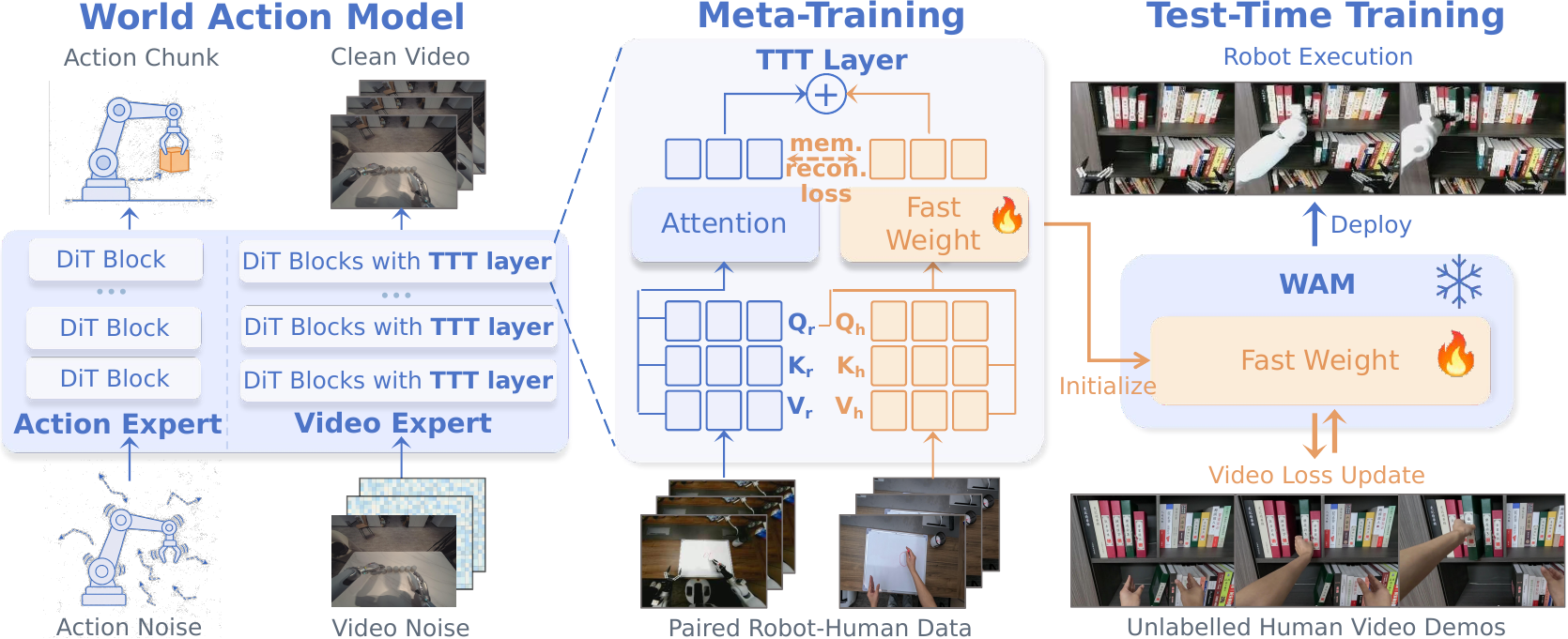}
    \caption{Pipeline of \ours. We first meta-train a fast-weight memory using paired human-robot demonstrations, encouraging human visual cues to align with robot behaviors through a key--value memory reconstruction objective. At test time, the memory is adapted from unlabeled human videos via video prediction, while the pretrained WAM remains frozen. The adapted memory then steers robot execution through the WAM's shared visual-action dynamics.}
	\label{fig:pipeline}
\end{figure}

\textbf{World Action Model.}
We build \ours on top of LDA~\citep{lda1b}, a pretrained world-action model (WAM). 
Each diffusion transformer block in LDA contains two coupled experts: a \emph{video expert} operating on visual latent tokens, and an \emph{action expert} operating on robot action tokens. 
The two experts communicate through joint attention. 
We denote the video and action tokens at block $\ell$ as $\bm{z}^{(\ell)}$ and $\bm{x}^{(\ell)}$, and the original LDA block output as $\hat{\bm{z}}^{(\ell+1)}$ and $\hat{\bm{x}}^{(\ell+1)}$.

\textbf{Video TTT layer.}
We keep the pretrained WAM architecture unchanged except for adding TTT residual branches to the video expert. 
For a TTT-augmented block, the output is
\begin{equation}
    \bm{z}^{(\ell+1)}
    =
    \hat{\bm{z}}^{(\ell+1)}
    +
    \Delta \bm{z}_{\mathrm{TTT}}^{(\ell)},
    \qquad
    \bm{x}^{(\ell+1)}
    =
    \hat{\bm{x}}^{(\ell+1)} .
    \label{eq:video_only_ttt}
\end{equation}
Each TTT layer follows the fast-weight memory formulation in prior TTT layers~\citep{sun2020ttt, wang2021tent, sun2025learnattesttime, zhang2026spatialttt, behrouz2025titans}. 
It contains slow projections $\theta_K^{(\ell)}, \theta_V^{(\ell)}, \theta_Q^{(\ell)}, \theta_O^{(\ell)}$ and a fast-weight network $f_{W^{(\ell)}}$. 
Given context tokens, the layer constructs Keys and Values; given the current video tokens, it constructs queries. 
After the fast weights are updated by the stage-specific TTT objective, the layer applies the fast-weight network to the video Queries:
\begin{equation}
    \Delta \bm{z}_{\mathrm{TTT}}^{(\ell)}
    =
    \theta_O^{(\ell)}
    f_{W^{(\ell)}}\!
    \left(
        \theta_Q^{(\ell)}(\bm{z}^{(\ell)})
    \right).
    \label{eq:ttt_residual}
\end{equation}

\subsection{Human-Robot Meta-Training}
\label{sec:method:midtrain}

\textbf{Meta-training objective.}
\hbc{Given a paired human-robot demonstration, $\bm{u}_h$ denotes the action-free human clip. At layer $\ell$, $\bm{h}_\phi^{(\ell)}$ and $\bm{z}_r^{(\ell)}$ are human and robot video tokens aligned at normalized phase $\phi=t/T_r$. The projections $\theta_{K,V,Q,O}^{(\ell)}$ and initialization $W_{\mathrm{init}}^{(\ell)}$ are slow parameters; $W_i^{(\ell)}$ denotes fast weights after $i$ inner updates. Human keys and values write into $f_W$, and robot queries read the adapted memory. The residual $\Delta\bm{z}_{\mathrm{TTT}}^{(\ell)}$ is added to the video stream and passed to later joint video--action blocks; it is not a direct residual on the action tokens.}
Since each TTT block's video output (Eq.~\ref{eq:video_only_ttt}--\ref{eq:ttt_residual}) is shifted by the residual \(\theta_O^{(\ell)} f_{W^{(\ell)}}(\theta_Q^{(\ell)}(\bm{z}^{(\ell)}))\), the fast weights \(\{W^{(\ell)}\}\) enter the per-block video stream and therefore propagate through to both (i) the final-block video latent \(\bm{z}^{(L)}\), on which the human video prediction loss \(\mathcal{L}_{\mathrm{vg}}^{\mathrm{human}}\) is computed, and (ii) the per-layer key--value memory reconstruction loss \(\mathcal{L}_{\mathrm{KVM}}^{(\ell)}\) defined below that probes how well \(f_W\) maps human Keys to human Values.
We adapt the fast weights on the combined inner-loop signal of these two objectives.

\textbf{Key--value memory reconstruction loss.} For each TTT layer \(\ell\), synchronized human tokens are projected into keys and values, while robot video tokens are projected into queries:
\[
    \bm{K}_h^{(\ell)}
    =
    \theta_K^{(\ell)}(\bm{h}_\phi^{(\ell)}),
    \qquad
    \bm{V}_h^{(\ell)}
    =
    \theta_V^{(\ell)}(\bm{h}_\phi^{(\ell)}),
    \qquad
    \bm{Q}_r^{(\ell)}
    =
    \theta_Q^{(\ell)}(\bm{z}_r^{(\ell)}).
\]
The per-layer memory reconstruction loss measures how well the current fast weights reconstruct the human values from the human keys:
\hbc{Here $B$, $L_h$, and $d$ denote batch size, human token count, and projected feature dimension.}
\begin{equation}
    \mathcal{L}_{\mathrm{KVM}}^{(\ell)}(W_i)
    =
    \frac{1}{B L_h d}
    \left\|
        f_{W_i^{(\ell)}}(\bm{K}_h^{(\ell)})
        -
        \bm{V}_h^{(\ell)}
    \right\|_2^2 .
    \label{eq:kv_aux}
\end{equation}

\textbf{Inner-loop adaptation.}
Propagating \(\bm{u}_h\) through the \(L\) TTT-augmented blocks via Eq.~\ref{eq:video_only_ttt}--\ref{eq:ttt_residual} produces the final video latents \(\bm{z}^{(L)}(\bm{u}_h;\,\Theta_{\mathrm{WAM}},\,\theta_{\mathrm{TTT}},\,\{W^{(\ell)}\})\), on which \(\mathcal{L}_{\mathrm{vg}}^{\mathrm{human}}\) is the standard LDA video-prediction loss; the \(W\)-dependence of \(\mathcal{L}_{\mathrm{adapt}}\) below is exactly this propagation.
Starting from \(W_0^{(\ell)}=W_{\mathrm{init}}^{(\ell)}\), the fast weights are updated by inner SGD on the combined human-side objective:
\begin{equation}
    \mathcal{L}_{\mathrm{adapt}}(W_i)
    =
    \mathcal{L}_{\mathrm{vg}}^{\mathrm{human}}
    \bigl(
        \bm{u}_h;\,
        \Theta_{\mathrm{WAM}},\,
        \theta_{\mathrm{TTT}},\,
        W_i
    \bigr)
    +
    \lambda \sum_{\ell} \mathcal{L}_{\mathrm{KVM}}^{(\ell)}(W_i),
\end{equation}
\begin{equation}
    W_{i+1}^{(\ell)}
    =
    W_i^{(\ell)}
    -
    \eta\,
    \nabla_{W_i^{(\ell)}}
    \mathcal{L}_{\mathrm{adapt}}(W_i),
    \label{eq:meta_inner_vg}
\end{equation}
where \(i\!\in\!\{0,1,\dots,N\}\) indexes the inner SGD iteration and \(\lambda\) weights the memory reconstruction term (see Table~\ref{tab:hp}).
The adapted weight \(W_N^{(\ell)}\) is what the residual readout in Eq.~\ref{eq:ttt_residual} uses.
Both terms in \(\mathcal{L}_{\mathrm{adapt}}\) depend only on the action-free human side, so the same inner-loop signal remains available at test time (Section~\ref{sec:method:tttdeploy}).

\textbf{Outer loss.}
The updated fast weights \(W_N^{(\ell)}\) are queried by \(\bm{Q}_r^{(\ell)}\) on the robot side and produce the residual in Eq.~\ref{eq:ttt_residual}.
The outer training objective is the standard WAM multitask loss on the paired robot data, which combines a video diffusion target on the robot video latents and an action diffusion target on the robot action chunks, inherited from the underlying LDA backbone~\citep{lda1b}:
\begin{equation}
    \mathcal{L}_{\mathrm{meta}}
    =
    \mathcal{L}_{\mathrm{WAM}}^{\mathrm{robot}}
    +\lambda\sum_\ell
      \mathcal{L}_{\mathrm{KVM}}^{(\ell)}(W_N) .
    \label{eq:meta_loss}
\end{equation}
\hbc{The outer objective also includes human key--value memory reconstruction at the adapted fast weights $W_N$, weighted by the same $\lambda$ as in the inner loop.}
Gradients are backpropagated through the TTT residual and the inner fast-weight update.
The optimized parameters are the WAM parameters, the TTT slow projections \(\theta_{\{K,V,Q,O\}}\), and the initialization \(W_{\mathrm{init}}\).
The adapted fast weights are discarded after each training example and reinitialized from \(W_{\mathrm{init}}\).

\textbf{Human-robot data synchronization.} To support training with alignment, we conduct offline sychronization for human-robot data pairs.
For a robot timestep $t$ in an episode of length $T_r$, we compute the normalized phase $\phi=t/T_r$ and select the nearest-phase frame from the paired human video of length $T_h$.

\subsection{Test-Time Training from Human Video}
\label{sec:method:tttdeploy}

At deployment, the WAM parameters, TTT slow projections, and \(W_{\mathrm{init}}\) are frozen.
The input to test-time training is a small batch of action-free human videos \(\mathcal{B}_h\) from the target domain.
We run the model in video-generation mode and optimize only the video-side TTT fast weights on the same combined objective form used at meta-training:
\begin{equation}
    \mathcal{L}_{\mathrm{TTT}}(W_i)
    =
    \frac{1}{|\mathcal{B}_h|}
    \sum_{\bm{u} \in \mathcal{B}_h}
    \!\left[
        \mathcal{L}_{\mathrm{vg}}
        \bigl(\bm{u};\, \Theta_{\mathrm{WAM}},\, \theta_{\mathrm{TTT}},\, W_i\bigr)
        +
        \lambda \sum_{\ell} \mathcal{L}_{\mathrm{KVM}}^{(\ell)}(\bm{u};\, W_i)
    \right],
    \label{eq:test_video_loss}
\end{equation}
\begin{equation}
    W_{i+1}^{(\ell)}
    =
    W_i^{(\ell)}
    -
    \eta\,
    \nabla_{W_i^{(\ell)}}
    \mathcal{L}_{\mathrm{TTT}}(W_i).
    \label{eq:test_update}
\end{equation}
Both \(\mathcal{L}_{\mathrm{vg}}\) and \(\mathcal{L}_{\mathrm{KVM}}^{(\ell)}\) are computed from the human side alone (Eq.~\ref{eq:kv_aux}), so no robot-side supervision is needed.
No WAM parameter, TTT slow projection, initialization parameter, or action-expert parameter is updated.
After \(N\) test-time updates (the same step budget as in Eq.~\ref{eq:meta_inner_vg}; see Table~\ref{tab:hp}), the adapted fast weights \(W_N\) are fixed during robot rollout:
\begin{equation}
    \bm{a}_{t:t+k}
    \sim
    p_{\Theta_{\mathrm{WAM}},\, \theta_{\mathrm{TTT}},\, W_N}
    \bigl(
        \bm{a}_{t:t+k}
        \mid
        \bm{o}_t,\, \bm{g}
    \bigr).
    \label{eq:rollout_policy}
\end{equation}


\section{Experiments}
\label{sec:experiments}

We evaluate \ours on real-robot manipulation across three embodiments. The protocol matches Section~\ref{sec:method}: a WAM is pre-trained, then the WAM is undergone human-robot meta training, and at deployment the TTT branch's fast weights adapt online via inner SGD on a small set of unseen-task human demonstrations while the WAM and slow weights stay frozen.

\subsection{Experimental Setup}
\label{sec:exp:setup}

\begin{figure}[t]
\centering
\includegraphics[width=0.95\linewidth]{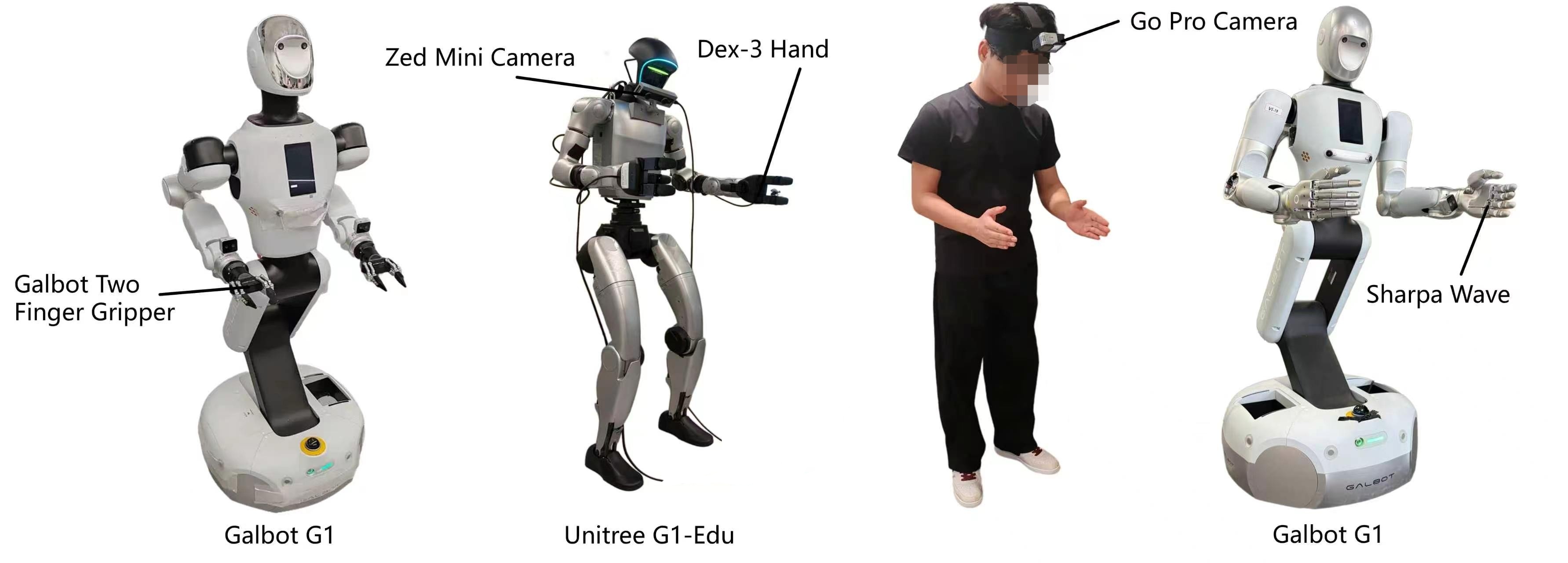}
\caption{Experimental setup.}
\label{fig:setup}
\end{figure}

\textbf{Robot and Tasks.} We evaluate \ours across three real-robot embodiments---\textbf{Unitree G1} (humanoid), \textbf{Galbot gripper} (bimanual two-finger), and \textbf{Galbot sharpa} (bimanual dexterous)---on a total of \textbf{9 manipulation tasks}: \emph{Transfer Bottle}, \emph{Table Bussing}, \emph{Deliver Drink}, \emph{Swap Place}, \emph{Pour Water}, \emph{Stamp Paper}, \emph{Flip Steak}, \emph{Pyramid Stacking}, and \emph{Multi-step Steak}. Each task is assigned to a single embodiment and is evaluated under two settings. The \emph{Orig.} setting collects evaluation trials inside the \textbf{standardized robot cubicle} that was also used to record the training data, with matching lighting, table height, and object instances. The \emph{New} setting deploys the robot in previously unseen \textbf{household environments} where lighting, table height, and the manipulated objects all change jointly relative to training---i.e., a combined out-of-distribution perturbation rather than a single-factor shift. We report \emph{progress} (\%) over 25 trials per (task, setting) cell. Progress is the standard partial-credit metric used in recent VLA evaluations: each trial receives 1.0 for full task completion and a fractional score in $[0,1]$ proportional to the number of pre-defined subgoals reached.


\textbf{Dataset and Metric.} We collect a meta-training dataset consisting of 2,286 paired human and robot episodes, which broadly covers 9 distinct manipulation tasks. Both robot and human data are captured from an egocentric perspective. Specifically, human demonstrations are recorded using a GoPro camera, without any form of pose estimation. 

\subsection{Compared with baselines}
\label{sec:exp:main}

\textbf{Baselines.} We compare against five baselines plus our \ours. \textsc{LDA}~\citep{lda1b}: the pretrained WAM backbone, no human data, no TTT branch. \textsc{WAM-Cotrain}: the same WAM further trained with paired human play data via the WAM multitask objectives (co-training; no TTT branch). \textsc{WAM-ICL}: the same WAM that ingests deployment-time human videos as in-context demonstrations, with no fast-weight adaptation. \textsc{EgoScale}~\citep{zheng2026egoscale}: a recent VLA scaled on diverse egocentric human data; as the original model is not open-source, we evaluate our re-implementation. \textsc{$\pi_{0.5}$}~\citep{intelligence2025pi05}: Physical Intelligence's open-world-generalization VLA. 

\begin{figure}[t]
  \centering
  \includegraphics[width=0.95\linewidth]{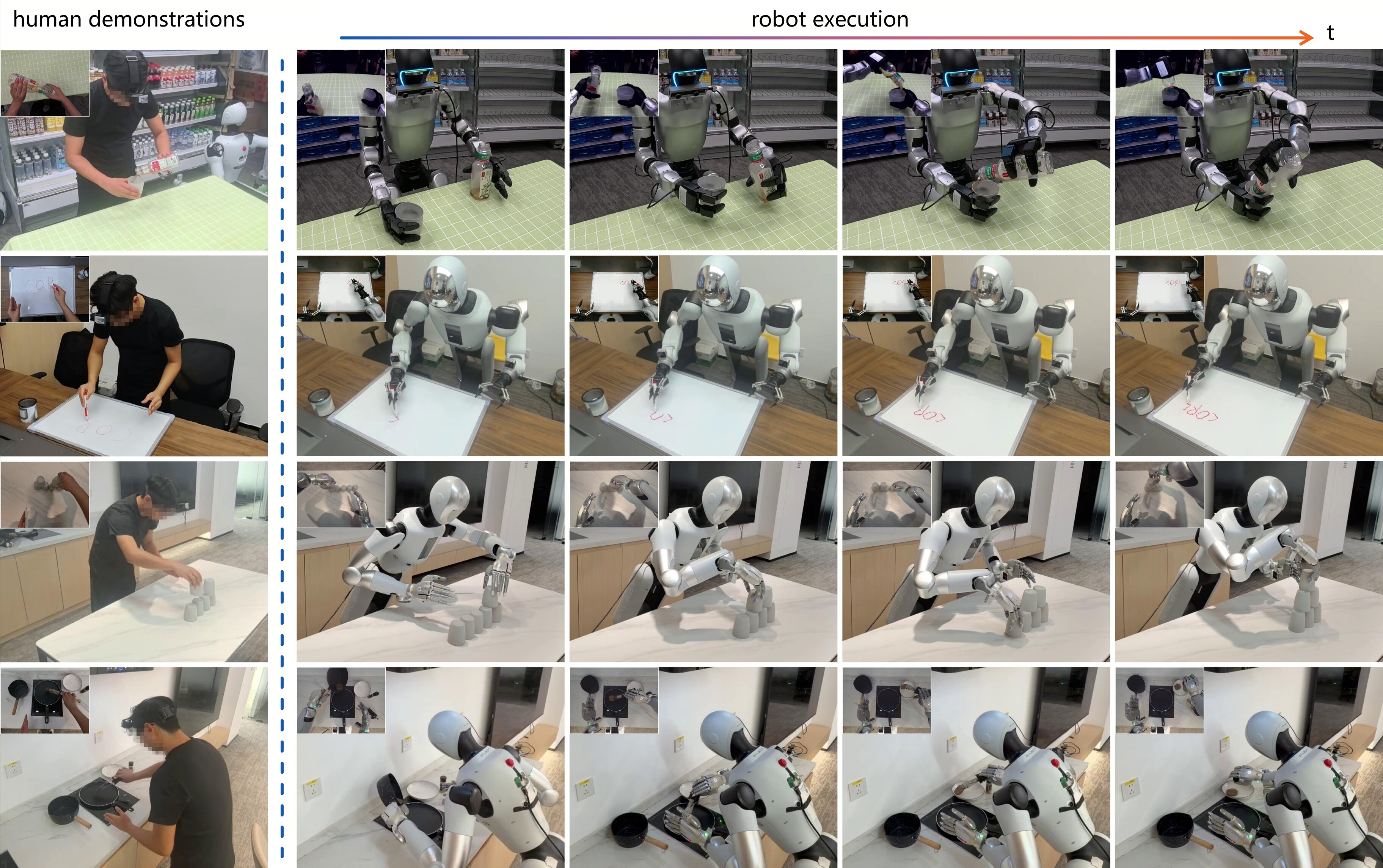}
  \caption{\textbf{Qualitative rollouts.} For each unseen task we show a robot rollout filmstrip (right) and the paired human demonstration used as deployment-time Key/Value (left).}
  \label{fig:gallery}
\end{figure}

\begin{table}[H]
\centering
\small
\caption{\textbf{Main results.} Progress (\%) on 9 manipulation tasks evaluated in previously unseen home environments. All cells averaged over 25 trials. The full table including the in-cubicle \emph{Orig.} setting is in Appendix~\ref{app:main-full}.}
\label{tab:main}
\setlength{\tabcolsep}{3.2pt}
\renewcommand{\arraystretch}{1.15}
\resizebox{\textwidth}{!}{
\begin{tabular}{lcccccccccc}
\toprule
Method
& \shortstack{Transfer \\Bottle}
& \shortstack{Table\\ Bussing}
& \shortstack{Deliver \\ Drink}
& \shortstack{Swap \\Place}
& \shortstack{Pour \\Water}
& \shortstack{Stamp \\ Paper}
& \shortstack{Flip\\ Steak}
& \shortstack{Pyramid \\ stacking}
& \shortstack{Multi-step \\Steak}
& Avg. \\
\midrule
\textsc{$\pi_{0.5}$}\cite{intelligence2025pi05}    & 33.4 & 36.0 & 15.0 &  7.4 & 10.0 & 24.4 &  2.0 &  4.7 &  0.3 & 14.8 \\
\textsc{LDA~\citep{lda1b}}            & 56.0 & 70.0 & 55.0 & 44.4 & 20.0 & \textbf{33.3} & 10.0 &  0.6 &  3.0 & 32.5 \\
\textsc{EgoScale}~\citep{zheng2026egoscale}       & 34.4 & 44.0 & 33.3 &  6.0 &  7.5 &  1.1 &  5.0 &  2.0 &  2.0 & 15.0 \\
\textsc{WAM-Cotrain}    & 10.0 & 10.0 & 44.3 & 48.1 & 24.0 & 21.7  & 34.2 & \textbf{12.0} & 23.8 & 25.3 \\
\textsc{WAM-ICL}        & 10.0 & 10.0 & 14.2 & 10.0 &  0.0 &  5.0 & 10.0 &  2.0 &  2.5 &  7.1 \\
\midrule
\textbf{\textsc{WAM-TTT}} & \textbf{55.6} & \textbf{100.0} & \textbf{66.7} & \textbf{66.7} & \textbf{30.0} & 8.3 & \textbf{34.3} & 10.4 & \textbf{43.8} & \textbf{46.2} \\
\bottomrule
\end{tabular}}
\end{table}

\textbf{Quantitative Results.} Table~\ref{tab:main} reports per-task progress in the \emph{New} household setting; the full table including the in-cubicle \emph{Orig.} numbers is deferred to Appendix~\ref{app:main-full}. \ours averages \textbf{46.2\%} across the 9 tasks, against 32.5\% for the no-TTT \textsc{LDA}\citep{lda1b} backbone (\textbf{+13.7} pts), 25.3\% for \textsc{WAM-Cotrain} (\textbf{+20.9} pts), 15.0\% for \textsc{EgoScale}\citep{zheng2026egoscale} (\textbf{+31.2} pts), 14.8\% for \textsc{$\pi_{0.5}$}\cite{intelligence2025pi05} (\textbf{+31.4} pts), and 7.1\% for \textsc{WAM-ICL} (\textbf{+39.1} pts). Three observations follow. (i) The gap against \textsc{WAM-ICL} is the strongest piece of evidence for the design hypothesis: feeding the same human videos as in-context tokens fails to transfer skill to unseen home environments, whereas absorbing them as fast-weight memory does. (ii) The gap against \textsc{LDA} (same WAM, no human data, no TTT) quantifies the contribution of human play data; the gap against \textsc{$\pi_{0.5}$} and \textsc{EgoScale} (no test-time human videos at all) quantifies the contribution of test-time adaptation itself. (iii) Across the 9 tasks \ours wins 7 outright and ties on Flip Steak (10.0); the single  exception is \emph{Stamp Paper} (8.3 vs.\ \textsc{LDA}'s 33.3), where the in-cubicle stamp pose is geometrically tight and the household-scene perturbation breaks an alignment that the human videos do not visibly correct.

\begin{hbcrevision}
\textbf{Comparison with stronger human-video interfaces.}
We compare human-video interfaces on unseen Sharpa \emph{Swap Place}, fixing the LDA/WAM backbone, robot and human data, prompts, injection blocks, and evaluation trials (Table~\ref{tab:human-interfaces}). These are interface variants within the same WAM, rather than full original systems. LingBot-VA 2.0-style ICL~\citep{zhang2026lingbotva2} reaches 50.0\%, below \ours (66.7\%). A possible reason for WAM-ICL's poor performance in Table~\ref{tab:main} is that our WAM was not pretrained to condition action denoising on full deployment-time human videos; learning this conditioning through meta-training alone may require more data. Likewise, XSkill-style SAT~\citep{xu2023xskill} and UniSkill-style ISD/FSD~\citep{kim2025uniskill} rely on pretrained skill latents; their meta-training-only insertion performs poorly here. \ours instead writes human-video context into fast weights.
\end{hbcrevision}

\begin{table}[H]
\centering
\small
\caption{\hbc{Human-video interface comparison on unseen Sharpa \emph{Swap Place}.}}
\label{tab:human-interfaces}
\begin{tabular}{llr}
\toprule
Method & Human-video use & Progress (\%) \\
\midrule
WAM (no adapt.) & None & 44.4 \\
LingBot-style ICL & Full context & 50.0 \\
XSkill-style (+SAT) & Skill + progress & 10.4 \\
UniSkill-style & ISD skill latent & 15.2 \\
\textbf{WAM-TTT} & Fast-weight write & \textbf{66.7} \\
\bottomrule
\end{tabular}
\end{table}

\textbf{Qualitative Results.}
\label{sec:exp:gallery}
Figure~\ref{fig:gallery} shows robot rollouts on three representative unseen tasks alongside the human demonstrations used as deployment-time Key/Value. Additional ablations (data-ratio sweep, model architecture) are in Appendix~\ref{app:abl-extra}.

\hbc{We also integrate our video-side TTT layers into a FastWAM-based implementation~\citep{yuan2026fastwam} and evaluate robot-video prediction on Pyramid stacking. Appendix~\ref{app:additional-wam} compares short-window predictions conditioned on current robot observations with the corresponding ground-truth frames.}

\subsection{Ablation Study}
\label{sec:exp:abl}

We conduct ablations to isolate the contribution of each component in \ours. 
Table~\ref{tab:abl-protocol} reports progress over $10$ trials on \textit{Table Bussing} and \textit{Swap Place}. 
The full WAM-TTT model combines human-robot meta-training, a key--value memory reconstruction objective, and test-time adaptation of the video-side TTT layers from human videos.
\textbf{WAM-LoRA} replaces the TTT fast-weight mechanism with a generic parameter-efficient adaptation baseline. 
\textbf{w/o Meta Training} removes the human-robot meta-training stage, so the TTT branch is not explicitly trained to align human Keys/Values with robot Queries. 
\textbf{w/o Memory Recon.} removes the inner key--value memory reconstruction loss, disabling the structured write mechanism into fast weights.
\textbf{w/o TTT} removes human-video adaptation entirely and evaluates the frozen WAM.

This ablation separates the effects of three design choices: using human videos at deployment, representing them through TTT fast weights, and meta-training the Q/K/V interface with paired human-robot data. 
The comparison between WAM-TTT and \textbf{w/o TTT} measures the value of test-time human-video adaptation. 
The comparison with \textbf{WAM-LoRA} tests whether the improvement comes specifically from the TTT memory structure rather than generic low-rank adaptation. 
The drops from \textbf{w/o Meta Training} and \textbf{w/o Memory Recon.} further quantify the importance of learning a human-to-robot memory interface before deployment.

\begin{table}[H]
\centering
\small
\caption{\textbf{Protocol ablation} on training and test-time inference choices. Progress(\%) on \textit{Table Bussing} and \textit{Swap Place} under the \textit{New} setting; 10 trials per cell.}
\label{tab:abl-protocol}

\setlength{\tabcolsep}{4pt}
\renewcommand{\arraystretch}{1.1}

\begin{tabular}{lccccc}
\toprule
Task 
& \textsc{WAM-TTT}
& WAM-LoRA
& w/o Meta Training
& w/o Memory Recon.
& w/o TTT\\
\midrule

Table Bussing
& \textbf{100.0}
&30.0
& 9.0
& 66.7
& 40.0\\

Swap Place
& \textbf{88.9}
& 0.0
& 0.0
& 72.0
& 74.1\\

\bottomrule
\end{tabular}
\end{table}

\begin{figure}[t]
\centering
\includegraphics[width=0.95\linewidth]{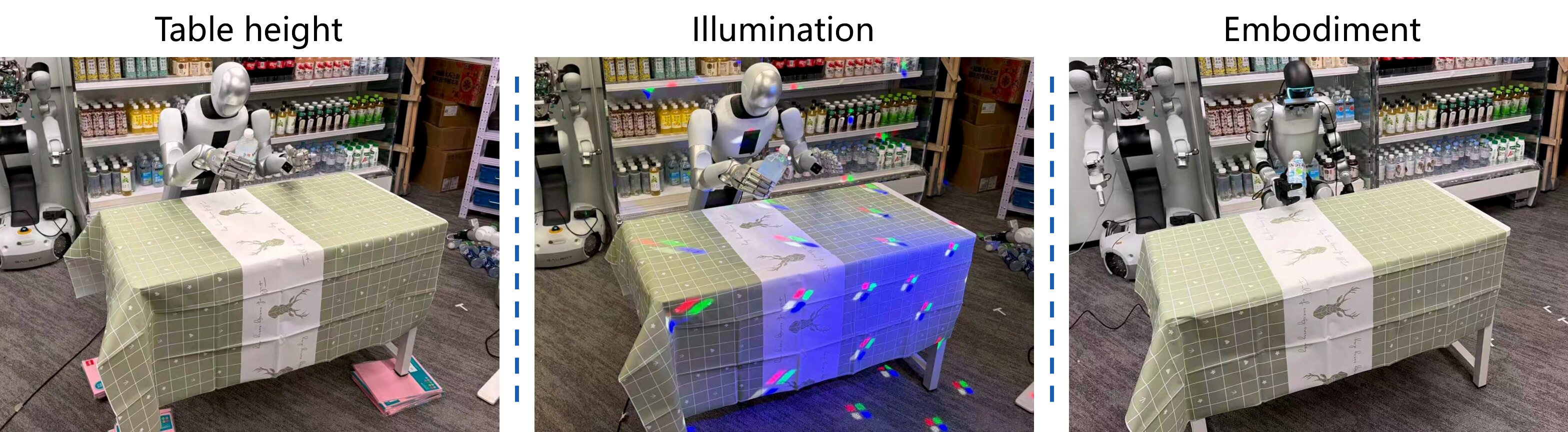}
\caption{Generalization Setup.}
\label{fig:generalization}
\end{figure}

\subsection{Generalization Preservation}
A potential concern is that test-time adaptation from a short human video may overfit the WAM to the demonstrated trajectory, sacrificing the broad generalization inherited from the pretrained foundation model. 
We evaluate this by first adapting \ours with human play videos from the target task and then testing the adapted policy under perturbations that change the execution condition, including lighting, object position, and embodiment-related appearance shifts. 
Table~\ref{tab:abl-gen} compares \ours with the pretrained LDA\citep{lda1b} backbone, a policy baseline $\pi_{0.5}$\citep{intelligence2025pi05} , and WAM-ICL, which uses the same human videos only as in-context demonstrations without fast-weight adaptation.

The key comparison is between WAM-ICL and WAM-TTT. 
Both methods receive the same human demonstration, but they use it in different ways: WAM-ICL conditions the frozen model on the demonstration at inference time, while \ours converts the human video into video-side fast weights through test-time training. 
Because the WAM backbone and action expert remain frozen, \ours steers the visual dynamics used for action generation without overwriting the pretrained action prior. 
As shown in Table~\ref{tab:abl-gen}, WAM-TTT maintains strong performance across all perturbation types on bimaual task \textit{Diliver Drink}. 
This indicates that the proposed TTT mechanism does not merely memorize the human demonstration; instead, it provides task- and domain-specific adaptation while preserving the foundation model's original robustness to visual and spatial shifts.

\begin{table}[t]
\centering
\small
\caption{\textbf{Evaluation of Generalization Ability.} }
\label{tab:abl-gen}

\setlength{\tabcolsep}{4pt}
\renewcommand{\arraystretch}{1.1}

\begin{tabular}{llcccc}
\toprule
Task & Perturbation 
& $\pi_{0.5}$\citep{intelligence2025pi05} 
& \textsc{LDA}\citep{lda1b} 
& \textsc{WAM-ICL} 
& \textbf{\textsc{WAM-TTT}} \\
\midrule

\multirow{2}{*}{Deliver Drink}
& Lighting
& 28.0 & 54.0 & 12.0 & \textbf{66.0} \\

& Spatial
& 0.0 & 28.0 & 20.0 & \textbf{56.0} \\




\bottomrule
\end{tabular}
\end{table}


\section{Conclusion, Limitation and Future Direction}
\label{sec:conclusion}

We presented \ours, a two-stage adaptation pipeline for World-Action Models. \emph{Human-robot meta-training} attaches Spatial-TTT-style\citep{zhang2026spatialttt} fast-weight branches to the WAM's video expert and jointly updates the WAM and the branches' slow projections via the WAM multitask outer loss, while the fast weights adapt online via inner SGD on self-supervised video-prediction and key--value memory reconstruction objectives derived from synchronized human--robot pairs. At \emph{test time}, the WAM, the action expert, and all slow projections are frozen; only the video-side fast weights update, via inner SGD on the user's unseen-task human videos. The recipe yields deployment-time skill absorption without any gradient step on the WAM, and matches or exceeds online-adaptation baselines on a real-robot manipulation suite (Section~\ref{sec:experiments}).

\paragraph{Limitations.} Three caveats. (1)~Meta-training phase pairing assumes the paired human episode covers the same skill phase distribution as the robot episode; mis-aligned manifests degrade the inner adaptation signal in a way the loss does not flag (Section~\ref{sec:exp:abl}). (2)~The deployment-time fast-weight adaptation is bounded by the expressiveness of the fast-weight network and by the slow projections fixed at meta-training; the further the deployment task drifts from the meta-training pairing distribution, the weaker the adaptation. We have not characterised the boundary empirically. (3)~Our deployment-time interface accepts only egocentric human RGB frames; it does not exploit hand-pose, contact, or 3-D scene cues that related work has shown useful~\citep{mu2026deximit,chen2025vidbot}.

\paragraph{Outlook.} The meta-training / test-time TTT interface generalises beyond human Key/Value: any auxiliary modality with a phase-pairable training signal could in principle drive a parallel fast-weight branch under the same loss-only-then-residual regime. We see \ours as a step toward WAM-based foundation model backbones whose attention structure carries explicit ``adaptation seats'' that downstream practitioners can drive with whatever side information they have at hand.


\clearpage
\acknowledgments{We thank the reviewers and the Area Chair for their careful review and constructive suggestions, which helped us improve the manuscript and strengthen the experiments.}


\bibliography{main}  

\appendix
\renewcommand{\theequation}{\thesection.\arabic{equation}}
\renewcommand{\thefigure}{\thesection.\arabic{figure}}
\renewcommand{\thetable}{\thesection.\arabic{table}}
\renewcommand{\thealgocf}{\thesection.\arabic{algocf}}
\makeatletter
\@addtoreset{equation}{section}
\@addtoreset{figure}{section}
\@addtoreset{table}{section}
\@addtoreset{algocf}{section}
\makeatother
\setcounter{equation}{0}
\setcounter{figure}{0}
\setcounter{table}{0}
\setcounter{algocf}{0}
\clearpage

\section{Meta-training algorithm}
\label{app:algo}

\paragraph{Notation: fast vs.\ slow weights.} Throughout, $W_{\mathrm{init}}^{(\ell)}$ denotes the learnable initialization of the fast-weight MLP at layer $\ell$ and is a \emph{slow} parameter trained by the outer optimizer; the projections $\theta_{\{K,V,Q,O\}}^{(\ell)}$ and the WAM parameters $\Theta_{\mathrm{WAM}}$ are likewise slow. The symbol $W_i^{(\ell)}$ (with $W_0^{(\ell)} \equiv W_{\mathrm{init}}^{(\ell)}$) denotes the \emph{fast} weights of layer $\ell$ at inner iteration $i$. The adjective ``fast'' refers to the time scale of updates per Spatial-TTT terminology~\citep{zhang2026spatialttt}: $W_i^{(\ell)}$ updates $N$ times \emph{within} a single forward pass via inner SGD, while $W_{\mathrm{init}}^{(\ell)}$ and the slow projections update only once per outer optimizer step (and are frozen at deployment).

\paragraph{Notation: token streams.} We use $\bm{z}^{(\ell)}$ for the video latent tokens and $\bm{x}^{(\ell)}$ for the robot action tokens at the input of LDA block $\ell$. These are the two streams that LDA's video expert and action expert process jointly via cross-stream attention (Section~\ref{sec:method:architecture}). The hatted symbols $\hat{\bm{z}}^{(\ell+1)}$ and $\hat{\bm{x}}^{(\ell+1)}$ denote the block's intrinsic outputs \emph{before} any TTT residual is added; the unhatted $\bm{z}^{(\ell+1)}, \bm{x}^{(\ell+1)}$ are what is actually fed into block $\ell+1$, which equals $\hat{\bm{z}}^{(\ell+1)} + \Delta\bm{z}_{\mathrm{TTT}}^{(\ell)}$ on the video stream and equals $\hat{\bm{x}}^{(\ell+1)}$ on the action stream (Eq.~\ref{eq:video_only_ttt}). The TTT residual modifies only the video stream, leaving the action expert's output untouched; this places test-time human-video adaptation entirely on the video side, in the modality where the action-free human videos can naturally supervise.

\paragraph{Notation: the outer-loop robot multitask loss \(\mathcal{L}_{\mathrm{WAM}}^{\mathrm{robot}}\).} We use $\mathcal{L}_{\mathrm{WAM}}^{\mathrm{robot}}$ as a shorthand for the WAM's standard multitask training loss evaluated on the paired robot side. It is the sum of two diffusion targets inherited verbatim from the LDA backbone~\citep{lda1b}: a video-side flow-matching / diffusion denoising loss on the robot video latents $\{\bm{z}_r^{(\ell)}\}$ output by the video expert, and an action-side flow-matching / diffusion denoising loss on the robot action chunks $\{\bm{x}_r^{(\ell)}\}$ output by the action expert. We adopt LDA's exact loss formulation, weighting, and noise schedule without modification; the only WAM-TTT contribution at this outer level is the TTT residual that shifts $\hat{\bm{z}}^{(\ell+1)}$ to $\bm{z}^{(\ell+1)}$ on the video stream and is then back-propagated through together with both diffusion targets.

\paragraph{The runtime target: cross-attention from robot queries to human keys/values.} At deployment, what we want each TTT layer to do is conceptually simple: let the robot token stream $\bm{z}_r$ \emph{read information from} the in-scene human token stream $\bm{h}$ through the standard attention interface. Letting $\bm{q} = \theta_Q(\bm{z}_r)$, $\bm{k}_i = \theta_K(\bm{h}_i)$, $\bm{v}_i = \theta_V(\bm{h}_i)$ collect the per-token projections of one query and of every human-side key/value, classical softmax cross-attention defines the desired readout
\begin{equation}
    \mathrm{Out}(\bm{q})
    \;=\;
    \sum_{i=1}^{L_h}\,
    \frac{\exp(\bm{q}^\top \bm{k}_i)}{\sum_j \exp(\bm{q}^\top \bm{k}_j)}\,
    \bm{v}_i .
    \label{eq:softmax_attention}
\end{equation}
Doing this literally would require materializing the full $(\bm{K}_h, \bm{V}_h)$ cache, whose length scales with the human-episode token count and which is awkward to re-update with each test-time gradient step.

\paragraph{The parametric substitute: an MLP that returns ``the value of the closest key''.} Instead of carrying the explicit cache, the TTT layer stores the human side inside the \emph{weights} $W$ of the fast-weight MLP $f_W$. The runtime readout is then the parametric expression already given by Eq.~\ref{eq:ttt_residual}, namely $\Delta\bm{z}_{\mathrm{TTT}} = \theta_O\,f_W\!\left(\theta_Q(\bm{z}_r)\right)$, with the slow $\theta_O$ projecting the $d$-dimensional output of $f_W$ back to the LDA hidden dimension so the residual can be added to $\hat{\bm{z}}^{(\ell+1)}$. The claim that this MLP-based readout behaves like cross-attention is purely a claim about the weights $W$: querying $f_W$ at $\bm{q}$ has to return the value associated with the key in the stored set that most resembles $\bm{q}$.

\paragraph{What makes $f_W$ act like attention: a linear-attention witness.} The deployed $f_W$ is a small nonlinear MLP, but the linear special case $f_W(\bm{x}) = W\bm{x}$ with $W \in \mathbb{R}^{d \times d}$ is already a tractable witness for what minimizing the key--value memory reconstruction loss $\mathcal{L}_{\mathrm{KVM}}$ does to the weights, and the nonlinear MLP is the smooth, normalized analog of the same retrieval pattern. Stack the human keys and values from the synchronized frame as $\bm{K}_h, \bm{V}_h \in \mathbb{R}^{L_h \times d}$ in the row-token convention of the notation paragraph above; the denominator $B L_h d$ in Eq.~\ref{eq:kv_aux} makes $\mathcal{L}_{\mathrm{KVM}}$ a per-element mean-squared error that is invariant to mini-batch size, human-sequence length, and embedding dimension, and is the form analyzed here. Minimizing the linear-case loss has a closed-form solution
\begin{equation}
    \min_{W \in \mathbb{R}^{d \times d}}\;\,
    \tfrac{1}{B L_h d}\,
    \bigl\|\bm{K}_h\,W^\top - \bm{V}_h\bigr\|_F^2
    \;\;\Longrightarrow\;\;
    W^* \;=\; \bm{V}_h^\top\,\bm{K}_h\,(\bm{K}_h^\top \bm{K}_h)^{-1} .
    \label{eq:lkv_linear_lsq}
\end{equation}
Under the standard linear-attention / modern-Hopfield isotropy hypothesis $\bm{K}_h^\top \bm{K}_h \approx (L_h/d)\,\bm{I}_d$, which holds for whitened or random-projection-style features and which serves here as a sanity check rather than a strict modeling assumption, the solution collapses to the Hebbian / outer-product memory
\begin{equation}
    W^* \;\propto\; \bm{V}_h^\top\,\bm{K}_h \;=\; \sum_{i=1}^{L_h} \bm{v}_i\,\bm{k}_i^\top .
    \label{eq:lkv_outer_product}
\end{equation}
Querying with the robot side $\bm{Q}_r = \theta_Q(\bm{z}_r)$ then yields
\begin{equation}
    f_{W^*}(\bm{Q}_r) \;=\; W^*\,\bm{Q}_r \;\propto\; \sum_{i=1}^{L_h}\,(\bm{k}_i^\top \bm{Q}_r)\,\bm{v}_i ,
    \label{eq:lkv_linear_attention_readout}
\end{equation}
which is exactly a kernel-free, softmax-free linear-attention readout against $(\bm{K}_h, \bm{V}_h)$~\citep{katharopoulos2020linear}. In other words, $\mathcal{L}_{\mathrm{KVM}}$ is not an auxiliary regularizer next to the cross-attention behaviour; in the linear case it is the variational definition of that behaviour. Equations~\eqref{eq:lkv_linear_lsq}--\eqref{eq:lkv_linear_attention_readout} hold as exact equalities only in this linear special case; the nonlinear MLP we actually deploy obeys the same training target $f_W(\bm{K}_h) \approx \bm{V}_h$, but the closed-form $W^* Q_r = \sum_i (\bm{k}_i^\top \bm{Q}_r)\bm{v}_i$ decomposition is replaced by the MLP's smooth, learned attention-like readout, in which the layer's nonlinearity and normalization play the role of the softmax kernel. The residual $\theta_O\,f_W(\bm{Q}_r)$ then injects the resulting human-derived value back into the video stream of Eq.~\ref{eq:video_only_ttt}.

\paragraph{Why the inner loop drives the fast weights to this witness.} The inner SGD step (Eq.~\ref{eq:meta_inner_vg}) directly minimizes $\mathcal{L}_{\mathrm{KVM}}$ alongside the human video-prediction loss $\mathcal{L}_{\mathrm{vg}}^{\mathrm{human}}$, so each inner update of the fast weights $W$ is, by construction, a gradient step toward a $W'$ for which the linear-attention witness above applies. \hbc{The outer loop optimizes the slow parameters $\Theta_{\mathrm{WAM}}, \theta_{\{K,V,Q,O\}}$, and $W_{\mathrm{init}}$ using the robot multitask loss plus $\lambda\sum_\ell\mathcal{L}_{\mathrm{KVM}}^{(\ell)}(W_N)$ (Eq.~\ref{eq:meta_loss}), backpropagating through the differentiable inner update~\citep{zhang2026spatialttt}.} There is therefore no indirection between ``do well on human video prediction'' and ``encode a human key--value memory'': both are simultaneously and explicitly part of the inner-loop signal that shapes $W$.

\paragraph{Why the witness is preserved at deployment.} The test-time inner loop (Eq.~\ref{eq:test_update}) optimizes the same combined objective form as meta-training: human video prediction plus per-layer key--value memory reconstruction. Since $\bm{K}_h^{(\ell)} = \theta_K^{(\ell)}(\bm{h}_\phi^{(\ell)})$ and $\bm{V}_h^{(\ell)} = \theta_V^{(\ell)}(\bm{h}_\phi^{(\ell)})$ are derivable from action-free human videos alone, no robot-side supervision is required to evaluate either term at deployment. The same inner SGD that produces the linear-attention witness during meta-training (Eqs.~\eqref{eq:lkv_linear_lsq}--\eqref{eq:lkv_linear_attention_readout}) therefore continues to produce it at deployment on $\mathcal{B}_h$, and the residual $\theta_O\,f_W(\theta_Q(\bm{z}_r))$ remains the human-key/value cross-attention readout that the meta-training stage promised.

\paragraph{Algorithm.} Algorithm~\ref{alg:meta} below realizes one meta-training step. Stage~(ii), test-time TTT (Section~\ref{sec:method:tttdeploy}), uses the same per-block forward and the same combined inner-loop objective form, but freezes the WAM, the TTT slow projections, and $W_{\mathrm{init}}$ while only the fast weights $W$ adapt via Eq.~\ref{eq:test_update}.

\begin{algorithm}[h]
\caption{One meta-training step on a paired robot--human batch.}
\label{alg:meta}
\KwIn{Paired batch $\mathcal{B}$ of robot trajectories with action-free human videos; LDA-based WAM~\citep{lda1b} with $L$ diffusion transformer blocks; TTT slow projections $\theta_{\{K,V,Q,O\}}^{(\ell)}$; fast-weight initializations $W_{\mathrm{init}}^{(\ell)}$; inner iterations $N$, inner LR $\eta$; memory reconstruction weight $\lambda$.}

\tcp{1. phase-aligned human-robot sync (Section~\ref{sec:method:midtrain})}
For each robot timestep $t$, pick the nearest-phase human frame $\bm{h}_\phi$\;

\tcp{2. inner adaptation: $N$ full-network SGD steps on the combined human-side objective}
$W^{(\ell)} \gets W_{\mathrm{init}}^{(\ell)}$ for all $\ell$\;
\For{$i = 1,\dots,N$}{
  Run a full WAM forward on $\bm{u}^h$ with current $\{W^{(\ell)}\}$ and TTT residuals (Eq.~\ref{eq:ttt_residual})\;
  Compute per-layer $\mathcal{L}_{\mathrm{KVM}}^{(\ell)}$ on the synchronized human frame (Eq.~\ref{eq:kv_aux})\;
  Assemble the inner-loop loss $\mathcal{L}_{\mathrm{adapt}} = \mathcal{L}_{\mathrm{vg}}^{\mathrm{human}} + \lambda \sum_\ell \mathcal{L}_{\mathrm{KVM}}^{(\ell)}$; backprop and update all layers simultaneously by Eq.~\ref{eq:meta_inner_vg}\;
}

\tcp{3. robot forward with adapted $W_N$}
\For{$\ell = 1,\dots,L$}{
  LDA block forward gives $(\hat{\bm{z}}^{(\ell+1)},\hat{\bm{x}}^{(\ell+1)})$\;
  Apply TTT residual to the video stream by Eq.~\ref{eq:ttt_residual} and Eq.~\ref{eq:video_only_ttt}\;
}

\tcp{4. outer loss and backprop}
\DontPrintSemicolon
\hbc{Re-evaluate human KVM reconstruction at $W_N$. Compute $\mathcal{L}_{\mathrm{meta}} = \mathcal{L}_{\mathrm{WAM}}^{\mathrm{robot}} + \lambda\sum_\ell\mathcal{L}_{\mathrm{KVM}}^{(\ell)}(W_N)$ (Eq.~\ref{eq:meta_loss}); backprop through the inner update into the WAM, $\theta_{\{K,V,Q,O\}}$, and $W_{\mathrm{init}}$; optimizer step}\;
\end{algorithm}

\section{Hyperparameters and datasets}
\label{app:setup}

\paragraph{Hyperparameters.} See Table~\ref{tab:hp}.

\begin{table}[h]
\centering\small
\caption{Hyperparameters for the main \ours runs.}
\label{tab:hp}
\begin{tabular}{lc}
\toprule
Setting & Value \\
\midrule
WAM backbone & \makecell[l]{LDA~\citep{lda1b} (Qwen3-VL-4B-Instruct VLM \\ \quad\,+ DiT-L MMDiT action head)} \\
MMDiT blocks $L$ / hidden dim $D$ / heads $H$ & 16 / 1536 / 32 \\
TTT head dim $d$ / fast-weight hidden width $f_h$ & 48 / 128 \\
Inner SGD iterations $N$ (meta-training and test-time) & 1 \\
Inner LR $\eta$ at meta-training & 0.1 \\
Inner LR $\eta$ at test time & 0.01 \\
Memory reconstruction weight $\lambda$ & $4\!\times\!10^{-2}$ \\
Outer optimizer & \makecell{AdamW ($\beta_1\!=\!0.9, \beta_2\!=\!0.999$), \\ weight decay $10^{-8}$} \\
Outer LR (DiT action head / VLM interface) & $1\!\times\!10^{-4}$ / $1\!\times\!10^{-5}$ \\
LR schedule & cosine with min $5\!\times\!10^{-7}$, 5\,k-step warmup \\
Meta-training steps & 100\,k \\
Batch size (per device / global) & 16 / 128 \\
GPUs & $8 \times$~NVIDIA H800, DeepSpeed ZeRO-2 \\
\bottomrule
\end{tabular}
\end{table}

\begin{figure}[H]
\centering
\includegraphics[width=0.95\linewidth]{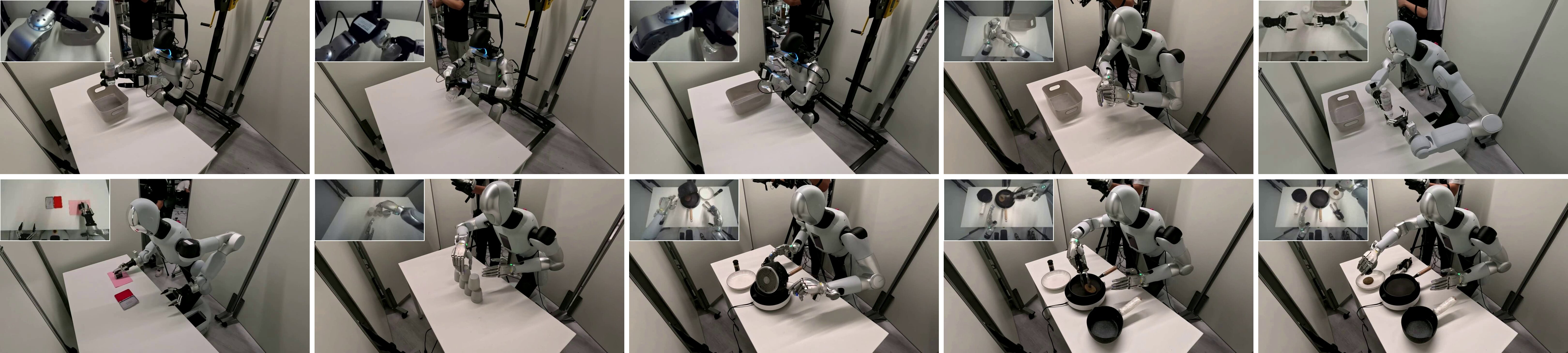}
\caption{\textbf{Robot data collection in the standardized cubicle.} Representative teleoperation frames in the $5\!\times\!2$ layout, read left-to-right, top-to-bottom. \textbf{Top row:} (1) \emph{Table Bussing}, (2) \emph{Pour Water}, (3) \emph{Deliver Drink} on the \textbf{Unitree~G1}; (4) \emph{Swap Place} on the \textbf{Galbot~sharpa}; (5) \emph{Transfer Bottle} on the \textbf{Galbot~gripper}. \textbf{Bottom row:} (6) \emph{Stamp Paper} on the \textbf{Galbot~gripper}; (7) \emph{Pyramid Stacking}, (8) \emph{Multi-step Steak} (grasping the pan and pouring the beef in), (9) \emph{Flip Steak}, (10) \emph{Multi-step Steak} (sprinkling pepper), all on the \textbf{Galbot~sharpa}. The long-horizon \emph{Multi-step Steak} occupies panels~8 and~10.}
\label{fig:data-robot}
\end{figure}

\begin{figure}[H]
\centering
\includegraphics[width=0.95\linewidth]{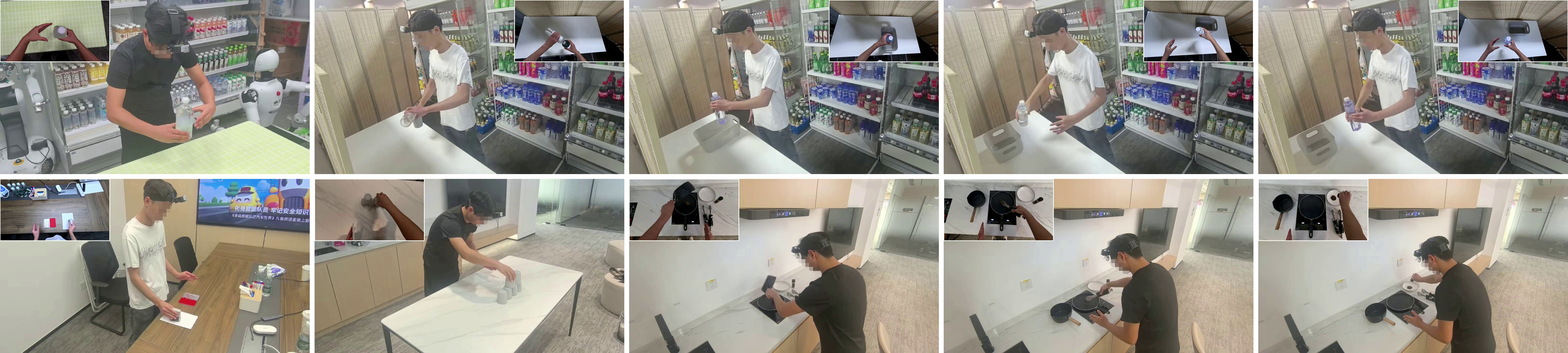}
\caption{\textbf{Human data collection in the actual New household scenes.} Same 10 task slots and panel order as Figure~\ref{fig:data-robot}, but the human demonstrator is in the actual household environment used as the \emph{New} evaluation setting (lighting, clutter, tableware, and target instances all differ from the cubicle). The hand pose varies per panel according to the paired robot end-effector: parallel-jaw mimic on the \textbf{Galbot~gripper} panels (5, 6), three-finger dex-hand grasp on the \textbf{Unitree~G1} panels (1--3), and unconstrained dexterous use on the \textbf{Galbot~sharpa} panels (4, 7--10). The task identity and panel order are the same as in Figure~\ref{fig:data-robot}. Demonstrations are recorded with a GoPro camera in egocentric view, without any hand-pose, joint-angle, or motion-retargeting annotation.}
\label{fig:data-human}
\end{figure}

\paragraph{Embodiments and datasets.} Three robot embodiments. \textbf{Unitree~G1} (humanoid bimanual, three-finger dex hand) covers \emph{Table Bussing}, \emph{Pour Water}, and \emph{Deliver Drink}. \textbf{Galbot~gripper} (bimanual two-finger) covers \emph{Transfer Bottle} and \emph{Stamp Paper}. \textbf{Galbot~sharpa} (bimanual dexterous, 22-DoF per side, 58-dim total) covers \emph{Swap Place}, \emph{Pyramid Stacking}, \emph{Flip Steak}, and the long-horizon \emph{Multi-step Steak}. Across all three embodiments we collect 2{,}286 paired robot-human episodes spanning these 9 manipulation tasks: 600 on Unitree~G1, 544 on Galbot~gripper, and 1{,}142 on Galbot~sharpa. Robot data is captured via teleoperation inside a standardized cubicle (Figure~\ref{fig:data-robot}), while the paired human demonstrations are recorded with a GoPro camera in egocentric view \emph{directly in the actual household environments} that we later evaluate as the \emph{New} setting (Figure~\ref{fig:data-human}), without any hand-pose, joint-angle, or motion-retargeting annotation. The two views are paired by phase alignment (Section~\ref{sec:method:midtrain}) for meta-training, and the human side is re-recorded in the deployment scene for test-time TTT (Section~\ref{sec:method:tttdeploy}). Each figure uses the same $5\!\times\!2$ grid (read left-to-right, top-to-bottom), with \emph{Multi-step Steak} occupying two panels (panels~8 and~10) so the 9 tasks fill the 10-cell layout.

\section{Full main results: Orig.\ and New settings}
\label{app:main-full}

Table~\ref{tab:main-full} is the full version of the main-paper result table (Table~\ref{tab:main} in Section~\ref{sec:exp:main}), including both the in-cubicle \emph{Orig.} setting and the unseen-household \emph{New} setting for each of the 9 manipulation tasks. The \emph{New} block is the one promoted to the main paper; the \emph{Orig.} block is provided here for completeness so the reader can verify how each baseline degrades under household-scene perturbation.

\begin{table*}[h]
\centering\small
\caption{\textbf{Full main results.} Progress (\%) on 9 tasks under the \emph{Orig.} (standardized robot cubicle) and \emph{New} (unseen household, combined OOD shift) settings. All cells averaged over 25 trials.}
\label{tab:main-full}
\setlength{\tabcolsep}{3pt}
\renewcommand{\arraystretch}{1.1}
\resizebox{\textwidth}{!}{
\begin{tabular}{llcccccccccc}
\toprule
Method & Setting
& \shortstack{Transfer \\ Bottle}
& \shortstack{Table \\ Bussing}
& \shortstack{Deliver \\ Drink}
& \shortstack{Swap \\ Place}
& \shortstack{Pour \\ Water}
& \shortstack{Stamp \\ Paper}
& \shortstack{Flip \\ Steak}
& \shortstack{Pyramid \\ Stacking}
& \shortstack{Multi-step \\ Steak}
& Avg. \\
\midrule
\multirow{2}{*}{\textsc{$\pi_{0.5}$}}
& Orig.  & 55.5 & 80.0 & 70.0 & 12.2 & 25.0 & 31.1 &  4.4 &  6.1 &  2.0 & 31.8 \\
& New    & 33.4 & 36.0 & 15.0 &  7.4 & 10.0 & 24.4 &  2.0 &  4.7 &  0.3 & 14.8 \\
\midrule
\multirow{2}{*}{\textsc{LDA}}
& Orig.  & 72.0 & 90.0 & 80.0 & 66.7 & 33.6 & \textbf{50.0} & 33.3 &  6.7 & 19.2 & 50.2 \\
& New    & \textbf{56.0} & 70.0 & 55.0 & 44.4 & 20.0 & \textbf{33.3} & 10.0 &  0.6 &  3.0 & 32.5 \\
\midrule
\multirow{2}{*}{\textsc{EgoScale}}
& Orig.  & 69.4 & 80.0 & 69.6 & 10.0 & 30.0 &  2.7 &  5.0 &  2.0 &  2.0 & 30.1 \\
& New    & 34.4 & 44.0 & 33.3 &  6.0 &  7.5 &  1.1 &  5.0 &  2.0 &  2.0 & 15.0 \\
\midrule
\multirow{2}{*}{\textsc{WAM-Cotrain}}
& Orig.  & 11.6 & 40.0 & 59.4 & 74.1 & 26.3 & 10.0 & 21.0 &  9.4 & 16.8 & 29.8 \\
& New    & 10.0 & 10.0 & 44.3 & 48.1 & 24.0 & 21.7 & 34.2 & \textbf{12.0} & 23.8 & 25.3 \\
\midrule
\multirow{2}{*}{\textsc{WAM-ICL}}
& Orig.  & 60.0 & 89.0 & 70.3 & 68.7 & 55.3 & 36.0 & 33.3 &  4.7 & 18.2 & 48.4 \\
& New    & 10.0 & 10.0 & 14.2 & 10.0 &  0.0 &  5.0 & 10.0 &  2.0 &  2.5 &  7.1 \\
\midrule
\multirow{2}{*}{\textbf{\textsc{WAM-TTT}}}
& Orig.  & \textbf{77.8} & \textbf{100.0} & \textbf{90.0} & \textbf{88.9} & \textbf{63.3} & 35.0 & \textbf{44.0} & \textbf{10.0} & \textbf{40.6} & \textbf{61.1} \\
& New    & 55.6 & \textbf{100.0} & \textbf{66.7} & \textbf{66.7} & \textbf{30.0} &  8.3 & \textbf{34.3} & 10.4 & \textbf{43.8} & \textbf{46.2} \\
\bottomrule
\end{tabular}}
\end{table*}

\paragraph{Analysis: \emph{Orig.}\ (standardized cubicle) results.} Under the \emph{Orig.} setting (top row of each block in Table~\ref{tab:main-full}), \ours leads at \textbf{61.1\%}, against 50.2\% for the no-TTT \textsc{LDA} backbone (\textbf{+10.9} pts), 48.4\% for \textsc{WAM-ICL} (\textbf{+12.7} pts), 31.8\% for \textsc{$\pi_{0.5}$} (\textbf{+29.3} pts), 30.1\% for \textsc{EgoScale} (\textbf{+31.0} pts), and 29.8\% for \textsc{WAM-Cotrain} (\textbf{+31.3} pts). Notably, \textsc{WAM-Cotrain}, which mixes paired human data into the WAM multitask outer loss without our TTT mechanism, drops \emph{below} the no-human \textsc{$\pi_{0.5}$} and \textsc{EgoScale} baselines: simply diluting robot supervision with human data without an explicit human-to-robot alignment mechanism actively damages in-distribution performance. \ours instead absorbs human data into a fast-weight memory that does not perturb the policy stream for unrelated robot trajectories, so it strictly improves over the WAM backbone in the same standardized setting.

\paragraph{Analysis: \emph{Orig.}\,$\to$\,\emph{New} transfer of human-data benefits.} The summary table below is computed directly from Table~\ref{tab:main-full}: the \emph{Orig.}\ and \emph{New} columns are the per-method 9-task averages (the \textbf{Avg} column of Table~\ref{tab:main-full}, one number per row of each \emph{Orig.}/\emph{New} block), and the retention ratio New/Orig and the gap $\Delta = \mathrm{Avg}_{\mathrm{Orig}} - \mathrm{Avg}_{\mathrm{New}}$ are derived from those two averages. The retention ratio thus measures how well each method's average standardized-cubicle competence survives the unseen-household perturbation, and $\Delta$ reports the average per-task progress lost in absolute points:
\begin{center}\small
\begin{tabular}{lcccc}
\toprule
Method & Orig. & New & New/Orig & $\Delta$ (Orig\,$-$\,New) \\
\midrule
\ours~(\textsc{WAM-TTT}) & 61.1 & 46.2 & \textbf{0.76} & $-14.9$ \\
\textsc{WAM-Cotrain}     & 29.8 & 25.3 & 0.85 & $-4.5$ \\
\textsc{LDA}             & 50.2 & 32.5 & 0.65 & $-17.7$ \\
\textsc{EgoScale}        & 30.1 & 15.0 & 0.50 & $-15.1$ \\
\textsc{$\pi_{0.5}$}     & 31.8 & 14.8 & 0.47 & $-17.0$ \\
\textsc{WAM-ICL}         & 48.4 &  7.1 & 0.15 & $-41.3$ \\
\bottomrule
\end{tabular}
\end{center}
\textsc{WAM-ICL}'s catastrophic collapse (15\% retention, $-41.3$ pts) shows that feeding human videos as in-context tokens is fragile under scene perturbation: the same long-context conditioning that helps in distribution becomes a liability when visual statistics shift. \textsc{WAM-Cotrain}'s high retention ratio (85\%) is largely artifactual because its starting point is already weak (29.8); in absolute terms it remains the second-worst method on \emph{New} after \textsc{WAM-ICL}. \ours combines the highest \emph{Orig.}\ performance (61.1\%) with the highest meaningful retention ratio (76\%), preserving most of the human-data benefit even when the deployment scene departs from the training cubicle.

\paragraph{Summary: form comparison among human-data methods.} Among the three forms of injecting paired human play data into a WAM, only \ours achieves the dual goal of strong in-distribution accuracy and robust OOD transfer: (i) direct multitask co-training (\textsc{WAM-Cotrain}) sacrifices in-distribution policy quality (lowest \emph{Orig.}\ score) and yields only modest \emph{New} gains; (ii) in-context conditioning (\textsc{WAM-ICL}) reaches reasonable \emph{Orig.}\ performance but collapses under household-scene shift; (iii) \ours's fast-weight TTT memory adapts the WAM only along the human-derived task evidence, leaving the LDA backbone's pretrained visual reasoning intact, so the human-data signal survives the OOD shift. This is the empirical signature of a useful human-data-injected memory: task-specific adaptation \emph{without} overwriting the WAM's transferable structure.

\section{Per-task progress definitions}
\label{app:task-progress}

We list the per-task subgoal decomposition used to compute the \emph{progress} (\%) reported in Tables~\ref{tab:main} and~\ref{tab:main-full}. Following the additive-scoring convention in recent VLA evaluations (e.g.,~\citep{zheng2026egoscale}), each task is decomposed into a small set of pre-defined milestones with weights summing to $1.0$; a trial's progress score is the sum of milestone weights reached, with $1.0$ reserved for trials that satisfy the final goal. Weights below are taken directly from our production evaluation rubric and are scored automatically from the robot's end-effector pose and known scene-object poses captured during the rollout. Progress is averaged across the 25 trials per (task, setting) cell.

\paragraph{Design rationale of the per-task weights.} Within each task we deliberately concentrate the bulk of the weight on the few critical milestones whose completion implies task success---e.g., \emph{Stamping successful} ($+0.50$) in Stamp Paper, \emph{Pouring successful} ($+0.60$) in Pour Water, \emph{Flipping successful} ($+0.45$) in Flip Steak, and \emph{Successfully place the 3rd cup} ($+0.36$) in Pyramid Stacking. This makes the metric reward primarily for finishing the task, in line with the binary success-rate interpretation a reader expects from a manipulation benchmark. We then assign small fractional weights to easier preliminary steps such as reaching toward an object before grasping it. These small weights extract useful signal from lower-quality trials that complete the early phases but stall later in the rollout, which we find informative for both behavioural-cloning training and downstream reinforcement-learning fine-tuning.

\paragraph{Transfer Bottle.} Instruction: \emph{``Hand the bottle from the left arm to the right arm and place it into the receiving box.''} Additive rubric:
\begin{itemize}
  \item $+0.05$: left hand reaches for the bottle.
  \item $+0.10$: left hand successfully grasps the bottle.
  \item $+0.05$: right hand reaches to receive the bottle.
  \item $+0.15$: right hand successfully grasps the bottle.
  \item $+0.10$: left hand releases the bottle.
  \item $+0.05$: right hand reaches to place the bottle.
  \item $+0.50$: right hand successfully places the bottle into the box.
\end{itemize}

\paragraph{Table Bussing.} Instruction: \emph{``Clear the tableware items from the table into the bin.''} Additive rubric, with $N$ items per trial (default $N\!=\!1$):
\begin{itemize}
  \item $+0.5/N$ per item: item grasped from the table.
  \item $+0.5/N$ per item: item released into the bin.
\end{itemize}

\paragraph{Deliver Drink.} Instruction: \emph{``Pick up the drink and hand it to a designated location.''} Additive rubric:
\begin{itemize}
  \item $+0.30$: drink (cup or bottle) grasped.
  \item $+0.30$: drink transported toward the recipient.
  \item $+0.40$: drink placed or released at the recipient.
\end{itemize}

\paragraph{Swap Place.} Instruction: \emph{``Pass the object from the left hand to the right hand, then place it at a designated location.''} Additive rubric:
\begin{itemize}
  \item $+0.20$: object A picked up.
  \item $+0.30$: object A staged in a buffer location.
  \item $+0.30$: object B picked up and placed at A's original position.
  \item $+0.20$: object A placed at B's original position.
\end{itemize}

\paragraph{Pour Water.} Instruction: \emph{``Pour water from the bottle into the cup.''} Additive rubric:
\begin{itemize}
  \item $+0.10$: cup successfully grasped.
  \item $+0.15$: bottle successfully grasped.
  \item $+0.05$: pouring posture reached.
  \item $+0.60$: pouring successful.
  \item $+0.10$: bottle successfully placed back.
\end{itemize}

\paragraph{Stamp Paper.} Instruction: \emph{``Stamp the paper at the marked location after applying the ink paste.''} Additive rubric:
\begin{itemize}
  \item $+0.05$: reach for the stamp.
  \item $+0.15$: stamp s[uccessfully grasped.
  \item $+0.05$: reach to the ink paste.
  \item $+0.20$: ink paste successfully applied.
  \item $+0.05$: reach to stamp the paper.
  \item $+0.50$: stamping successful.
\end{itemize}

\paragraph{Flip Steak.} Instruction: \emph{``Use the tongs to flip the steak in the pan.''} Additive rubric:
\begin{itemize}
  \item $+0.02$: reach for the tongs.
  \item $+0.20$: tongs successfully grasped.
  \item $+0.03$: approach the steak.
  \item $+0.20$: steak successfully clamped.
  \item $+0.45$: flipping successful.
  \item $+0.10$: tongs successfully put down.
\end{itemize}

\paragraph{Pyramid Stacking.} Instruction: \emph{``Stack six cups into a three-layer pyramid (a base layer of three cups, a middle layer of two, and a single top cup).''} Additive rubric (each ``1st/2nd/3rd cup'' entry below tracks the layer-defining placement of that layer):
\begin{itemize}
  \item $+0.01$: reach for the 1st cup.
  \item $+0.05$: 1st cup successfully grasped.
  \item $+0.02$: reach to place the 1st cup.
  \item $+0.10$: 1st cup successfully placed.
  \item $+0.01$: reach for the 2nd cup.
  \item $+0.10$: 2nd cup successfully grasped.
  \item $+0.02$: reach to place the 2nd cup.
  \item $+0.15$: 2nd cup successfully placed.
  \item $+0.01$: reach for the 3rd cup.
  \item $+0.15$: 3rd cup successfully grasped.
  \item $+0.02$: reach to place the 3rd cup.
  \item $+0.36$: 3rd cup successfully placed.
\end{itemize}

\paragraph{Multi-step Steak.} Instruction: \emph{``Plate the steak from the pan, flip it during cooking, transfer it back to the plate, and season it with pepper.''} Additive rubric:
\begin{itemize}
  \item $+0.01$: reach for the pan.
  \item $+0.02$: pan successfully held.
  \item $+0.01$: reach to place the steak.
  \item $+0.07$: steak successfully poured into the pan.
  \item $+0.01$: reach to place the pan.
  \item $+0.01$: pan placed successfully.
  \item $+0.01$: reach for the tongs.
  \item $+0.07$: tongs successfully held.
  \item $+0.01$: reach to clamp the steak.
  \item $+0.17$: steak successfully clamped.
  \item $+0.20$: steak flipping successful.
  \item $+0.15$: steak successfully clamped again.
  \item $+0.07$: steak successfully placed onto the plate.
  \item $+0.01$: reach to place the tongs.
  \item $+0.01$: tongs put down.
  \item $+0.01$: reach for the pepper bottle.
  \item $+0.07$: pepper bottle successfully held.
  \item $+0.09$: sprinkling pepper successful.
\end{itemize}

\section{Additional results}
\label{app:abl-extra}

\subsection{Qualitative rollout gallery in unseen household, office, and kitchen scenes}
\label{app:gallery-new}

\begin{figure}[H]
\centering
\includegraphics[width=0.95\linewidth]{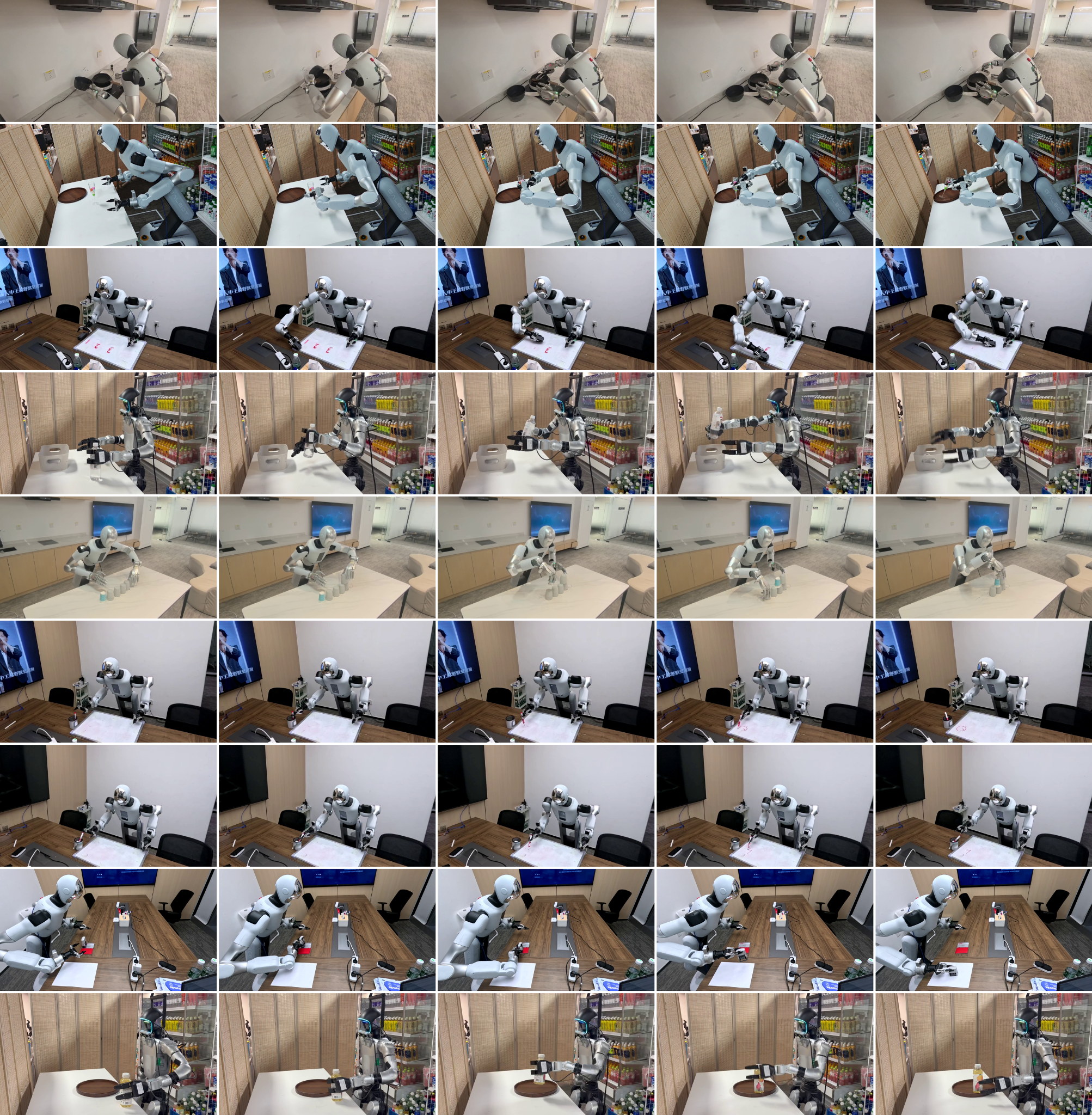}
\caption{\textbf{Qualitative gallery of \ours~rollouts in unseen household, office, and kitchen scenes.} Each row is a single rollout of one task, shown as 5 evenly-spaced keyframes (left $\to$ right, initial $\to$ terminal frame). \textbf{Rows from top to bottom:} (1) \emph{Multi-step Steak} in a completely new kitchen with the stovetop raised $+10$\,cm relative to the meta-training cubicle (Galbot~sharpa); (2) \emph{Transfer Bottle} with a novel long-stem wine-glass instance (Galbot~gripper); (3) \emph{Wipe Blackboard} in a meeting room (Galbot~gripper); (4) \emph{Deliver Drink} (Unitree~G1, dex-3 hand); (5) \emph{Pyramid Stacking} on the Galbot~sharpa, with the leftmost cup replaced by a novel paper-cup instance mid-task; (6) \emph{Draw a circle} on the meeting-room whiteboard (Galbot~gripper); (7) \emph{Stamp Paper} in the meeting room with the target stamp position shifted away from the cubicle anchor (Galbot~gripper); (8) free-form handwriting of the letter \emph{``L''} (Galbot~gripper); (9) \emph{Table Bussing} (Unitree~G1, dex-3 hand). Rows (1, 2, 4, 5, 7, 9) draw from the 9-task benchmark of Section~\ref{sec:exp:setup} under the \emph{New} perturbation, while rows (3, 6, 8) are additional in-the-wild demonstrations beyond the headline benchmark.}
\label{fig:gallery-new}
\end{figure}

Figure~\ref{fig:gallery-new} compiles 9 representative \ours~rollouts in unseen scenes. Six of the rows correspond to evaluated tasks from the 9-task benchmark of Section~\ref{sec:exp:setup} executed under the \emph{New} setting (Multi-step Steak, Transfer Bottle, Deliver Drink, Pyramid Stacking, Stamp Paper, Table Bussing); the remaining three rows are additional in-the-wild demonstrations exercising perturbation axes that the benchmark does not test (whiteboard wiping, free-form circle drawing, and handwriting), included to convey the breadth of skills the model retains after meta-training and test-time TTT. Each row is a single rollout of one task, shown as 5 evenly-spaced keyframes read left-to-right, with the leftmost column showing the initial observation and the rightmost column showing the final scene at episode termination. All rollouts come from the same checkpoint used to produce Table~\ref{tab:main-full}, after test-time TTT adaptation from in-scene human videos (Section~\ref{sec:method:tttdeploy}); no per-task hyperparameter sweep is performed.

\subsection{Direct lab-scene generalization without scene-specific human data}
\label{app:lab-gen}

We further stress-test \ours's deployment-time generalization by moving the robot into a previously unseen lab scene while keeping the model exactly as it left meta-training. Crucially, \emph{no additional in-scene human videos} are collected for this evaluation: the TTT branch only uses the slow projections and $W_{\mathrm{init}}$ shaped by paired robot-human play during meta-training. This isolates the contribution of the meta-trained slow projections from any test-time human-video adaptation in the deployment scene.

\begin{figure}[H]
\centering
\includegraphics[width=0.95\linewidth]{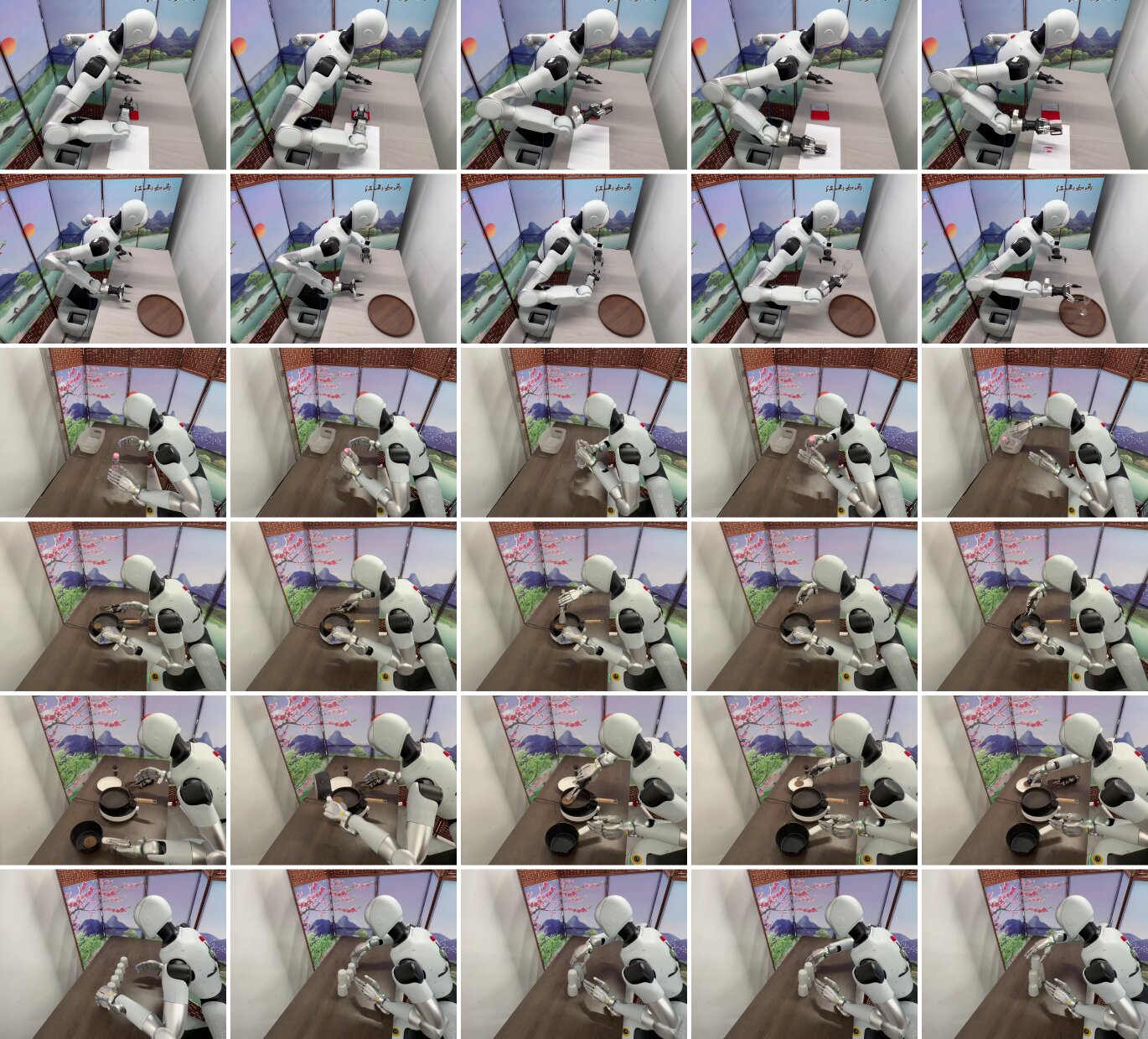}
\caption{\textbf{Direct generalization to an unseen lab scene without scene-specific human data.} \ours~rollouts after meta-training only; no in-scene human videos are provided, so the TTT branch is driven entirely by the meta-trained slow projections and $W_{\mathrm{init}}^{(\ell)}$. The lab scene differs from the standardized cubicle in lighting, tablecloth, object instance, and target-pose offset \emph{simultaneously}. From top to bottom, the six rollout strips show: \emph{Stamp Paper} and \emph{Transfer Bottle} on the \textbf{Galbot gripper}; \emph{Swap Place}, \emph{Flip Steak}, \emph{Multi-step Steak}, and \emph{Pyramid Stacking} on the \textbf{Galbot sharpa}. The behaviour transfers despite the joint OOD shift, showing that the calibration achieved during meta-training is not tied to the in-cubicle observation statistics.}
\label{fig:lab-gen}
\end{figure}

Figure~\ref{fig:lab-gen} reports rollouts on \textbf{Galbot gripper} and \textbf{Galbot sharpa} across four perturbation axes simultaneously relative to the standardized robot cubicle: (i) lighting (warm vs.\ cool, side- vs.\ top-mounted), (ii) tablecloth pattern and colour, (iii) novel object instances of the same category, and (iv) target-pose offsets. Despite all four perturbations being applied jointly, the WAM together with the meta-trained slow projections retains a useful skill prior in the new lab scene, indicating that the calibration achieved during meta-training is not tied to the in-cubicle observation statistics. This direct-transfer result complements Table~\ref{tab:main-full}: it shows that even without populating the Human-Key/Value cache from in-scene demonstrations, \ours can recover useful behaviour purely from the WAM-side knowledge that the meta-training stage has shaped.

\subsection{Robustness across six axes of in-scene distribution shift}
\label{app:six-axis-gen}

Appendix~\ref{app:lab-gen} stress-tested \ours with \emph{no} in-scene human videos at all. Here we consider the complementary stress test: the deployment scene \emph{does} provide in-scene human videos, but the scene itself (and therefore the human videos used by test-time TTT) departs from the meta-training cubicle along six different perturbation axes that we evaluate one at a time. Each axis is applied to both the robot rollout scene and the paired human-side videos consumed by the test-time inner loop (Eq.~\ref{eq:test_update}), so each axis is a single-factor OOD perturbation of the entire deployment signal rather than a synthetic perturbation of only one stream.

This is the regime where standard adaptation pipelines tend to fail. Methods that \emph{co-train} on the perturbed human data (e.g.\ the \textsc{WAM-Cotrain} baseline of Section~\ref{sec:exp:setup}) propagate the human-side domain shift into the full backbone, so a few perturbed in-scene videos are enough to noticeably erode the pretrained policy. Methods that condition on human videos \emph{in context} (e.g.\ \textsc{WAM-ICL}) likewise feed the perturbed observation statistics directly into the conditioning stream, which Table~\ref{tab:main-full} already shows collapses sharply on the household \emph{New} split for the same reason. \ours's TTT branch avoids both failure modes by construction: at test time only the fast weights $W^{(\ell)}$ are updated, while $\Theta_{\mathrm{WAM}}$, the slow projections $\theta_{\{K,V,Q,O\}}^{(\ell)}$, and the fast-weight initialization $W_{\mathrm{init}}^{(\ell)}$ all remain frozen (Eq.~\ref{eq:test_update}). The human-side domain shift can therefore only rewrite the human-to-robot memory; it cannot rewrite the policy itself, and the WAM's pretrained visual reasoning and action prior are preserved.

\begin{figure}[H]
\centering
\includegraphics[width=0.95\linewidth]{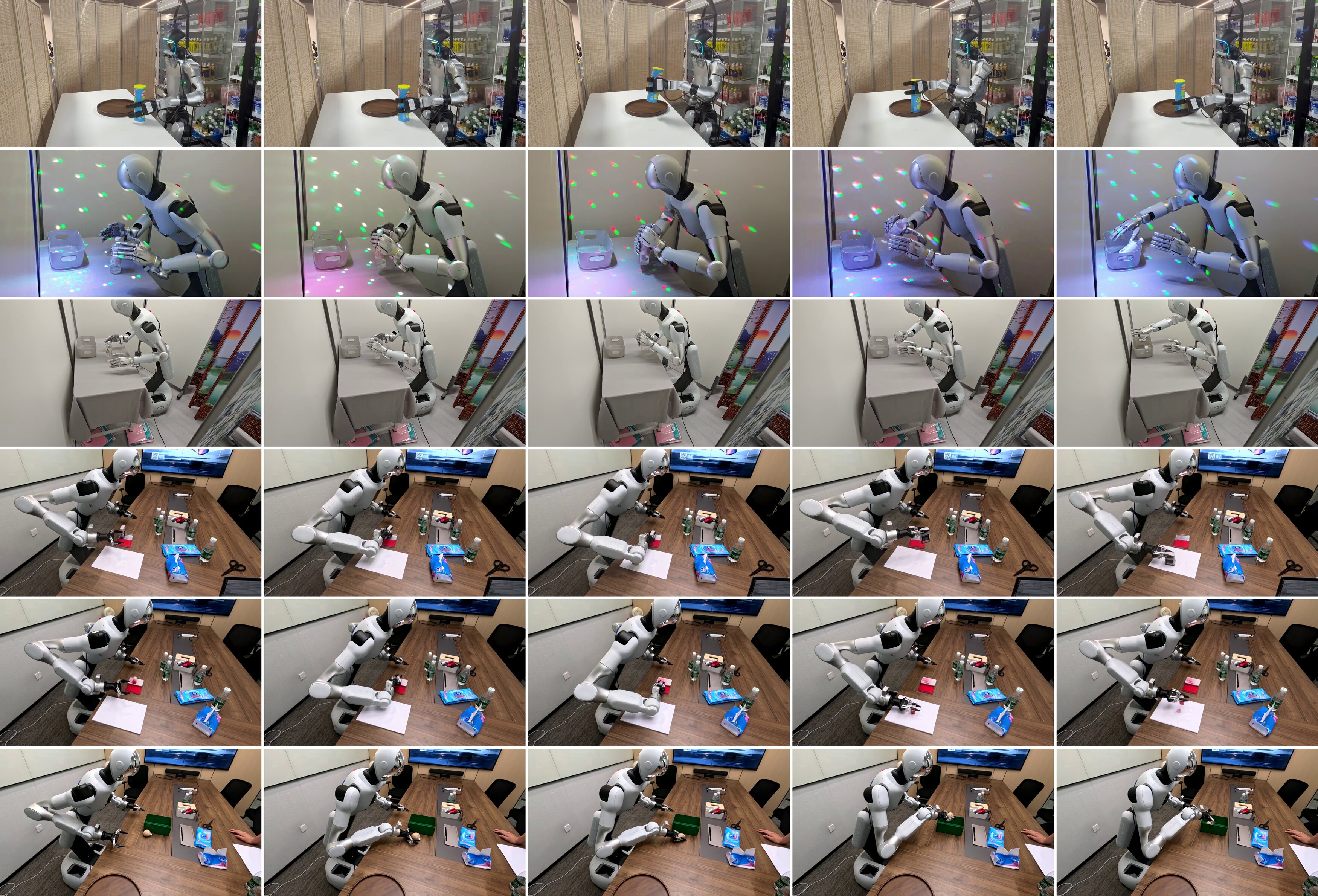}
\caption{\textbf{\ours~across six axes of in-scene distribution shift.} Each row is a single rollout under one perturbation axis, shown as 5 evenly-spaced keyframes (left $\to$ right, initial $\to$ terminal frame). \textbf{Rows from top to bottom:} (1) \emph{object} generalization, novel object instances on \emph{Table Bussing} (Unitree~G1, dex-3 hand); (2) \emph{lighting} generalization, heavy color-temperature / intensity / direction shift on \emph{Swap Place} (Galbot~sharpa); (3) \emph{height} generalization, raised-table on \emph{Swap Place} (Galbot~sharpa); (4) \emph{chassis} generalization, chassis position shifted plus cluttered desk on \emph{Stamp Paper} (Galbot~gripper); (5) \emph{brand-new stamp} / affordance shift on \emph{Stamp Paper} (Galbot~gripper); (6) \emph{object-position} generalization, target placement outside the meta-training distribution (Galbot~gripper). All rollouts come from the same checkpoint as Table~\ref{tab:main-full}; the test-time inner loop is fed the perturbed in-scene human videos with no per-axis hyperparameter tuning.}
\label{fig:six-axis-gen}
\end{figure}

The six perturbation axes shown in Figure~\ref{fig:six-axis-gen}, in row order top-to-bottom:
\begin{itemize}
    \item \textbf{Object generalization} (Unitree~G1, dex-3 hand, \emph{Table Bussing}). The tabletop is populated with novel object instances drawn from categories that the meta-training set never includes.
    \item \textbf{Lighting generalization} (Galbot~sharpa, \emph{Swap Place}). The dominant light source is changed in color temperature (warm $\leftrightarrow$ cool), intensity, and direction relative to the cubicle.
    \item \textbf{Height generalization} (Galbot~sharpa, \emph{Swap Place}). The table is raised relative to the standardized cubicle height, shifting the robot-to-target geometry.
    \item \textbf{Chassis (background) generalization} (Galbot~gripper, \emph{Stamp Paper}). The robot chassis is repositioned and the desk is cluttered with distractor items, perturbing the background statistics of the egocentric scene.
    \item \textbf{Brand-new stamp / affordance generalization} (Galbot~gripper, \emph{Stamp Paper}). The stamp is replaced by an unfamiliar instance whose shape and grasp affordance differ from any stamp seen at training.
    \item \textbf{Object-position generalization} (Galbot~gripper). The target placement lies outside the meta-training distribution, requiring the policy to interpolate a manipulation pose it has never been demonstrated.
\end{itemize}
All six rows are produced from the same checkpoint used to fill Table~\ref{tab:main-full}; the test-time inner loop is run on the perturbed in-scene human videos for that axis, with no checkpoint switching or per-axis hyperparameter tuning. For the full rollout videos at native frame rate, please refer to our accompanying supplementary video.

This result is the architectural reason \ours is robust where vanilla adaptation is not: residual fast-weight TTT decouples the human-side update from the WAM-side parameters, so domain shift in the human videos cannot reach into and overwrite the WAM's pretrained capability.

\subsection{Data-ratio ablation}
\label{app:abl-data}

Our default meta-training budget is $(r,h) = (100,100)$ paired robot/human episodes per task. Table~\ref{tab:abl-data} sweeps this ratio across three representative tasks. Rather than a full grid sweep, we keep one row per distinct question we want to answer:
\begin{itemize}
    \item \textbf{(100, 0)} -- no-human baseline at our robot budget; isolates the marginal value of paired human data.
    \item \textbf{Iso-budget triple at total $=200$ episodes:} \textbf{(200, 0)}, \textbf{(100, 100)}, and \textbf{(10, 190)}. The three rows hold the total data-collection cost fixed and only vary the robot/human split, so any difference between them is attributable to the mix rather than to scale. \textbf{(200, 0)} is the robot-only upper bound and answers the natural challenge ``why not just collect more robot data?''. \textbf{(10, 190)} pushes the mix to the cheap-human extreme, testing whether human data can carry the policy when robot teleoperation is the bottleneck resource. \textbf{(100, 100)} is the deployed configuration used everywhere else in the paper.
    \item \textbf{(100, 200)} -- adds 100 more paired human episodes on top of our default while keeping robot count fixed; probes whether the human-side gain has saturated at $h=100$.
\end{itemize}

\begin{table}[h]
\centering\small
\caption{\textbf{Data-ratio ablation.} Progress (\%) under the \emph{New} setting at varying meta-training data budgets. $r$ = robot demos per task, $h$ = paired human demos per task. Our deployed setup is $(100,100)$; each other row isolates a single question. All cells averaged over \textbf{25 trials per (configuration, task)}, matching the main-paper protocol of Section~\ref{sec:exp:setup}.}
\label{tab:abl-data}
\begin{tabular}{lcccc}
\toprule
(robot, human) & Transfer Bottle & Table Bussing & Deliver Drink & Avg. \\
\midrule
(100, 0)                          & 44.1          & 90.0           & 44.4          & 59.5 \\
\midrule
\multicolumn{5}{l}{\emph{Iso-budget triple: total $= 200$ episodes per task}} \\
(10, 190)                         & 42.1          & 68.0           & 44.2          & 51.4 \\
(100, 100) \emph{(ours)}          & 55.6          & \textbf{100.0} & 66.7          & \textbf{74.1} \\
(200, 0)                          & 47.9          & \textbf{100.0} & \textbf{73.2} & 73.7 \\
\midrule
(100, 200)                        & \textbf{58.9} & \textbf{100.0} & 61.0          & 73.3 \\
\bottomrule
\end{tabular}
\end{table}

\paragraph{Takeaways.} Three reads of the table line up with the design intent.
\emph{(i) Paired human data has a clear marginal value at fixed robot budget:} adding 100 paired human episodes on top of $(r,h)=(100,0)$ raises the 3-task average from 59.5 to 74.1 ($+14.6$ pts), and is essentially free relative to the robot-side cost.
\emph{(ii) At the same total budget of 200 episodes per task, paired human data substitutes 1-for-1 for robot data:} the iso-budget triple shows $(100,100)$ at $74.1$ and $(200,0)$ at $73.7$ are statistically indistinguishable on the 3-task average, so the practitioner can halve the robot teleoperation cost without sacrificing performance. The cheap-human extreme $(10,190)$ at $51.4$, however, falls well below both, confirming that some robot grounding is required and that human data is a substitute, not a replacement, for the action-conditioned signal.
\emph{(iii) The human side has already saturated at our default:} doubling human episodes to $(100,200)$ gives $73.3$, marginally below $(100,100)$ on the same 3-task average, so we use $h=100$ as the deployed setup.

\subsection{Model-architecture ablation}
\label{app:abl-model}

Whereas the main-paper baselines (Section~\ref{sec:exp:setup}) contrast \ours against alternative \emph{training recipes} that all share a VLM-conditioned DiT backbone, Table~\ref{tab:abl-model} instead ablates the \emph{backbone composition} itself: every row carries our full meta-training and test-time TTT pipeline, and only the VLM side is varied. The goal is to justify the architectural choice of a pretrained, fully unfrozen VLM as the conditioning backend for the DiT.

\begin{table}[h]
\centering\small
\caption{\textbf{Model-architecture ablation.} Progress (\%) on \emph{Table Bussing} under the \emph{New} setting. All variants retain the full meta-training plus test-time TTT pipeline of Section~\ref{sec:method}; only the VLM conditioning backend is changed. Each row averaged over \textbf{10 trials} per configuration; the main-paper 25-trial protocol of Section~\ref{sec:exp:setup} is reduced here to keep the four-way architectural sweep tractable, so single-row figures should be read with a wider error margin than in the data-ratio ablation of Table~\ref{tab:abl-data}.}
\label{tab:abl-model}
\begin{tabular}{lc}
\toprule
Configuration & Progress (\%) \\
\midrule
DiT only (no VLM backend)                                            & 72.0 \\
DiT + VLM (no VLM pretrain)                                          & 80.0 \\
DiT + VLM (VLM frozen)                                               & 54.0 \\
DiT + VLM (VLM open) \emph{(ours)}                                   & \textbf{100.0} \\
\bottomrule
\end{tabular}
\end{table}

\paragraph{Takeaways.} Three reads of the table justify the architectural choice. \emph{(i) The VLM backbone is load-bearing:} removing it entirely costs $-28$ pts (DiT-only at 72 vs.~ours at 100). \emph{(ii) The VLM's pretraining is load-bearing:} replacing the pretrained VLM with a randomly initialized one costs $-20$ pts (no-pretrain at 80 vs.~ours at 100), so the visual-language prior is doing real work beyond contributing capacity. \emph{(iii) Joint adaptation is not optional:} freezing the pretrained VLM costs $-46$ pts and drops below even the no-VLM configuration (54 vs.~72), indicating that a fixed pretrained representation, however general, becomes a bottleneck for the DiT once human-robot alignment requires adapting the conditioning features themselves. The joint meta-training of VLM and DiT is therefore the right setup for our pipeline.

\subsection{Action-pseudolabel ablation}
\label{app:abl-action-pseudo}

The original \ours~design treats human videos as \emph{action-free}: \hbc{at test time the combined objective $\mathcal{L}_{\mathrm{vg}}^{\mathrm{human}}+\lambda\sum_\ell\mathcal{L}_{\mathrm{KVM}}^{(\ell)}$ drives inner SGD on $W^{(\ell)}$ (Eq.~\ref{eq:test_video_loss}); both terms use human videos without robot action labels.} A natural alternative, used by several human-data pipelines~\citep{kareer2024egomimic,zheng2026egoscale}, is to extract a \emph{pseudo-action} for each human frame and then train an action-conditioned objective on the human side as well. We test this alternative here.

\begin{figure}[H]
\centering
\includegraphics[width=0.95\linewidth]{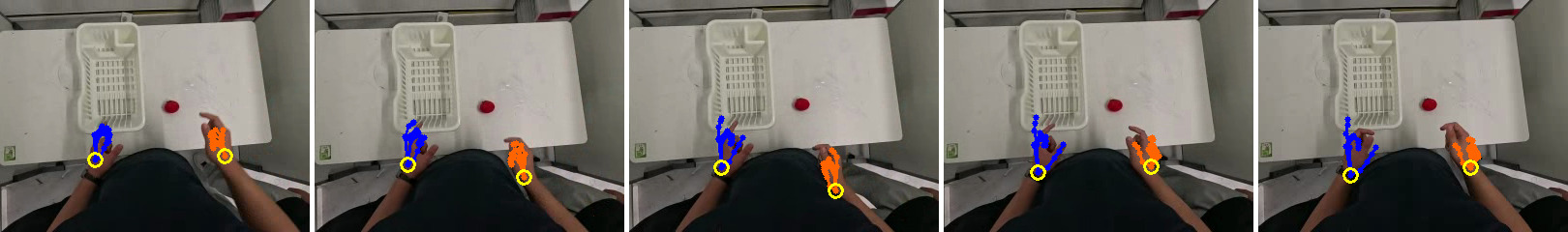}
\caption{\textbf{Representative single-view hand-pose annotation from the FD pipeline.} Five evenly-spaced frames of one human episode, overlaid with the MediaPipe~\citep{lugaresi2019mediapipe} keypoints and the fitted MANO~\citep{romero2017mano} mesh. The overlay is the input to the EgoScale-style~\citep{zheng2026egoscale} retargeter that produces the pseudo-qpos $\tilde{\bm{a}}_t^{\mathrm{human}}$ used by the \textsc{VG + FD} variant in Table~\ref{tab:abl-action-pseudo}. Visible imperfections of the monocular single-view fit (finger-tip drift, occluded thumb estimates, inconsistent palm normal across frames) propagate downstream into the retargeted pseudo-action and are the underlying reason the FD loss hurts.}
\label{fig:annotated-row}
\end{figure}

\paragraph{Pipeline.} For each human episode collected on the GoPro, we estimate the wrist 6-DoF pose with the MediaPipe hand tracker~\citep{lugaresi2019mediapipe} combined with the EgoMimic-style estimation pipeline~\citep{kareer2024egomimic}, fit a parametric MANO hand model~\citep{romero2017mano} to recover the full hand pose, and retarget the resulting fingertip and palm targets to the target embodiment's joint configuration using the optimization-based retargeting protocol of EgoScale~\citep{zheng2026egoscale}. The output is a sequence of pseudo-qpos $\tilde{\bm{a}}_t^{\mathrm{human}}$, the human-side analogue of robot teleop actions, available at the same frame rate as the egocentric video. Figure~\ref{fig:annotated-row} shows a representative 5-frame strip with the MediaPipe keypoints and MANO mesh overlaid on the original egocentric video, illustrating the typical quality of the single-view annotation that the FD pipeline consumes.

\paragraph{Objective added on the human side.} Given pseudo-actions, we can train one of the WAM-side objectives that the original \ours~drops on human data: a \emph{forward-dynamics} (FD) loss that, conditioned on the current observation $\bm{o}_t^{\mathrm{human}}$ and the pseudo-action $\tilde{\bm{a}}_t^{\mathrm{human}}$, predicts the next observation in the frozen DINOv3~\citep{simeoni2025dinov3} feature space. \hbc{The added prediction term is $\lambda_{\mathrm{FD}}\mathcal{L}_{\mathrm{FD}}^{\mathrm{human}}$. The combined human-side objective is $\mathcal{L}_{\mathrm{vg}}^{\mathrm{human}}+\lambda\sum_\ell\mathcal{L}_{\mathrm{KVM}}^{(\ell)}+\lambda_{\mathrm{FD}}\mathcal{L}_{\mathrm{FD}}^{\mathrm{human}}$;} we use $\lambda_{\mathrm{FD}} = 1$ throughout this ablation, so the FD term enters at the same scale as the video-generation term and is not artificially down-weighted. \hbc{At test time VG and KVM use the in-scene human videos, while the added FD term also uses their MANO-derived pseudo-actions.}

\paragraph{Comparison.} The two configurations compared in Table~\ref{tab:abl-action-pseudo} are: (i) \textsc{VG only (ours)}, the deployed action-free \ours~design, \hbc{where the human-side adaptation objective is $\mathcal{L}_{\mathrm{vg}}^{\mathrm{human}}+\lambda\sum_\ell\mathcal{L}_{\mathrm{KVM}}^{(\ell)}$, with no MANO retargeting or pseudo-action;} and (ii) \textsc{VG + FD (pseudo-action)}, the same backbone, the same meta-training schedule, the same paired robot-human dataset, and the same test-time TTT pipeline, but with the MANO retargeting pipeline (MediaPipe wrist + EgoMimic-style estimation $\to$ MANO hand $\to$ EgoScale-style retargeter $\to$ pseudo-qpos $\tilde{\bm{a}}_t^{\mathrm{human}}$ matched to the target embodiment) producing pseudo-actions, and with the DINOv3-feature-space FD loss added to the human side at $\lambda_{\mathrm{FD}} = 1$. The four tasks span all three embodiments and three end-effector families: \emph{Transfer Bottle} on the \textbf{Galbot gripper} (two-finger), \emph{Table Bussing} and \emph{Deliver Drink} on the \textbf{Unitree~G1} (dex-3 hand), and \emph{Swap Place} on the \textbf{Galbot sharpa} (22-DoF dexterous). Both configurations are evaluated under the \emph{New} household setting using the same main-paper 25-trial protocol of Section~\ref{sec:exp:setup}; the \textsc{VG only} row reproduces the WAM-TTT entry of Table~\ref{tab:main-full} for these four tasks, so the cross-row delta directly isolates the contribution of the MANO retargeting and the FD loss.

\begin{table}[h]
\centering\small
\caption{\textbf{Action-pseudolabel ablation.} Progress (\%) under the \emph{New} setting; \textbf{25 trials per (configuration, task)}. Per-task embodiment: Transfer Bottle = Galbot gripper, Table Bussing / Deliver Drink = Unitree G1 (dex-3 hand), Swap Place = Galbot sharpa (dexterous). \textsc{VG only (ours)} is the deployed action-free \ours~design (no MANO, no pseudo-action). \textsc{VG + FD (pseudo-action)} adds the MANO retargeting pipeline (MediaPipe~\citep{lugaresi2019mediapipe} + MANO~\citep{romero2017mano} + EgoScale-style retargeter~\citep{zheng2026egoscale}) to produce an embodiment-matched pseudo-qpos $\tilde{\bm{a}}_t^{\mathrm{human}}$ per human frame, and adds the DINOv3~\citep{simeoni2025dinov3}-feature-space forward-dynamics loss $\mathcal{L}_{\mathrm{FD}}^{\mathrm{human}}$ on the human side at $\lambda_{\mathrm{FD}} = 1$.}
\label{tab:abl-action-pseudo}
\begin{tabular}{lccccc}
\toprule
Configuration & Transfer Bottle & Table Bussing & Deliver Drink & Swap Place & Avg. \\
\midrule
\textsc{VG only} \emph{(ours)}    & \textbf{55.6} & \textbf{100.0} & \textbf{66.7} & \textbf{66.7} & \textbf{72.3} \\
\textsc{VG + FD} (pseudo-action) & 14.2          & 33.3           & 26.8          & 41.2          & 28.9 \\
\bottomrule
\end{tabular}
\end{table}

\paragraph{Takeaways.} Adding the retargeted pseudo-action and FD loss is uniformly harmful, dropping the 4-task average from 72.3 to 28.9 ($-43.4$ pts). The damage is largest on the two end-effector families where the MANO output does not map cleanly to the robot's actuation: $-41.4$ on \emph{Transfer Bottle} (Galbot gripper, a one-DoF parallel jaw), $-66.7$ on \emph{Table Bussing} and $-39.9$ on \emph{Deliver Drink} (Unitree G1 dex-3 hand). For both the gripper and the dex-3 hand, the binary or near-binary closure command is not naturally present in the MANO pose, so a hand-engineered post-processor is needed to derive the open/close signal. This post-processor compounds on top of the already-noisy single-view monocular hand-pose estimate, and the resulting pseudo-action is too far from the true robot action distribution to provide a useful FD supervision signal. Even on \emph{Swap Place} on the Galbot sharpa, where the dexterous robot is the most direct geometric target for the MANO output, FD still costs $-25.5$ pts ($66.7 \to 41.2$): the residual single-view retargeting noise alone is enough to corrupt the learned forward dynamics. The conclusion supports the design choice of \ours: under current single-view hand-tracking and retargeting maturity, injecting retargeted pseudo-actions into the human-side training signal is net-negative, and keeping human videos action-free is the right call.

\clearpage
\subsection{Video prediction with an additional WAM backbone}
\label{app:additional-wam}

\begin{hbcrevision}
We integrate our video-side TTT layers into a FastWAM-based implementation~\citep{yuan2026fastwam} and evaluate robot-video prediction on Pyramid stacking. Each window contains the current observation and eight predicted frames at 0.4\,s intervals, spanning 3.2\,s.
\end{hbcrevision}

\begin{figure}[H]
\centering
\includegraphics[width=\linewidth]{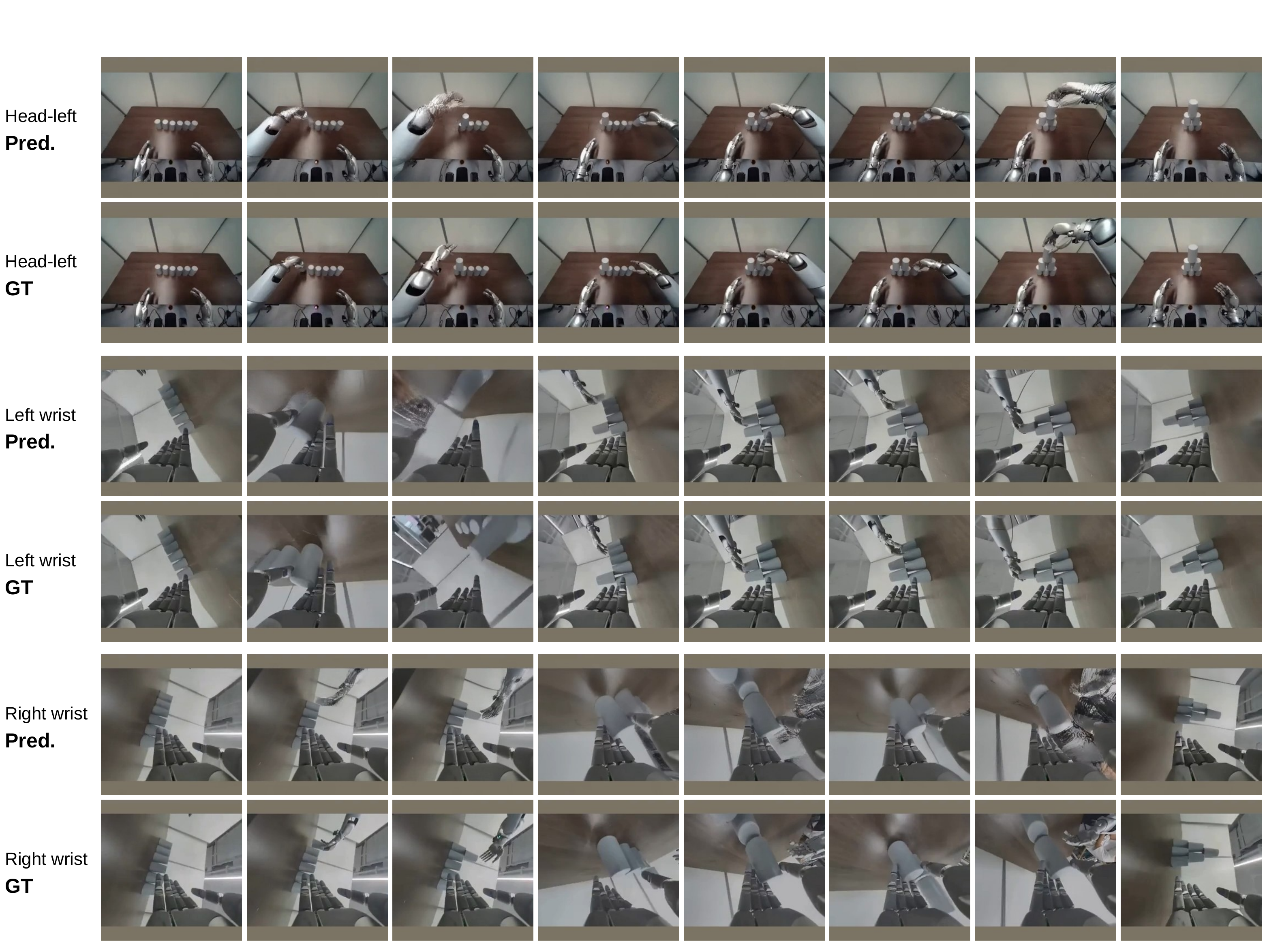}
\caption{\hbc{\textbf{Pyramid stacking.} Each pair of rows shows prediction (top) and ground truth (bottom) for the head-left, left-wrist, and right-wrist cameras, respectively.}}
\label{fig:fastwam-cups}
\end{figure}

\end{document}